\documentclass{article}

\PassOptionsToPackage{numbers, sort}{natbib}
 \usepackage[preprint]{neurips_2026}

\usepackage[utf8]{inputenc} 
\usepackage[T1]{fontenc}    
\usepackage{url}            
\usepackage{booktabs}       
\usepackage{amsfonts}       
\usepackage{nicefrac}       
\usepackage{microtype}      
\usepackage{xcolor}         
\usepackage{microtype}
\usepackage{graphicx}
\usepackage{amsmath}
\usepackage{amsthm}

\usepackage{amssymb}
\usepackage{amsfonts}
\usepackage{nicefrac}       
\usepackage{mathrsfs}
\usepackage{mathtools}
\usepackage{tabularx}
\usepackage{array}          
\usepackage{booktabs}
\usepackage{longtable}
\usepackage{wrapfig}
\usepackage{rotating}
\usepackage{multirow}
\usepackage{caption}
\usepackage{lipsum}
\usepackage[dvipsnames]{xcolor}
\usepackage{enumitem}
\usepackage{inconsolata}
\usepackage{pifont}
\usepackage{soul}
\usepackage{siunitx}
\usepackage{csquotes}
\usepackage{xspace}
\usepackage{comment}
\usepackage{adjustbox}
\usepackage{fontawesome}
\usepackage{marvosym}
\usepackage{pifont}
\usepackage{colortbl}
\usepackage{arydshln}
\usepackage{epigraph}
\usepackage[framemethod=TikZ]{mdframed}
\usepackage{enumitem}
\usepackage[normalem]{ulem}
\usepackage[accsupp]{axessibility}  
\usepackage{titletoc}
\usepackage{tikz,lipsum} 
\usepackage[most]{tcolorbox}
\usepackage[backref=page]{hyperref}
\usepackage[capitalize]{cleveref}
\usepackage{soul}
\usepackage{makecell}

\makeatletter
\def\thanks#1{\protected@xdef\@thanks{\@thanks
\protect\footnotetext{#1}}}
\makeatother

\definecolor{imageOrange}{RGB}{242, 169, 59}
\definecolor{textGreen}{RGB}{65, 124, 92}
\definecolor{instructionRed}{RGB}{195, 98, 95}
\definecolor{previousPink}{RGB}{176, 47, 162}
\definecolor{cgtGreen}{RGB}{146, 160, 62}
\definecolor{imstartBlue}{RGB}{96, 147, 212}

\newcommand{\imageOrange}[1]{\textcolor{imageOrange}{#1}}
\newcommand{\textGreen}[1]{\textcolor{textGreen}{#1}}
\newcommand{\instructionRed}[1]{\textcolor{instructionRed}{#1}}
\newcommand{\previousPink}[1]{\textcolor{previousPink}{#1}}
\newcommand{\cgtGreen}[1]{\textcolor{cgtGreen}{#1}}
\newcommand{\imstartBlue}[1]{\textcolor{imstartBlue}{#1}}

\renewcommand{\emph}[1]{\textit{#1}}
\renewcommand{\paragraph}[1]{\noindent\textbf{#1}}

\newcommand{\miniscule}{\fontsize{6}{7}\selectfont}

\newcommand{\qvl}{Qwen2.5-VL}
\newcommand{\lov}{LLaVA-OneVision}
\newcommand{\qvlshort}{QVL}
\newcommand{\lovshort}{LOV}

\newcommand{\reducenum}[1]{\textcolor{red}{{\miniscule $\downarrow$#1}}}
\newcommand{\increasenum}[1]{\textcolor{OliveGreen}{{\miniscule $\uparrow$#1}}}
\newcommand{\similarnum}[1]{\textcolor{gray}{{\miniscule $\sim$#1}}}

\newcommand{\bA}{\mathbf{A}}

\newtcolorbox{resultbox}{
colback=gray!5,
colframe=gray!50,
boxrule=0.4pt,
arc=2pt,
left=4pt,
right=4pt,
top=3pt,
bottom=3pt,
before skip=6pt,
after skip=6pt
}

\newtcolorbox{inlinebox}{
colback=gray!10,
colframe=gray!50,
boxrule=0.3pt,
arc=1.5pt,
left=3pt,
right=3pt,
top=2pt,
bottom=2pt,
before skip=4pt,
after skip=4pt
}

\crefname{figure}{Fig.}{Figs.}
\Crefname{figure}{Figure}{Figures}
\crefname{section}{Sec.}{Secs.}
\Crefname{section}{Section}{Sections}
\crefname{appendix}{App.}{Apps.}
\Crefname{appendix}{Appendix}{Appendices}
\crefname{table}{Tab.}{Tabs.}
\Crefname{table}{Table}{Tables}

\theoremstyle{plain}

\theoremstyle{definition}

\theoremstyle{remark}

\makeatletter
\DeclareRobustCommand\onedot{\futurelet\@let@token\@onedot}
\def\@onedot{\ifx\@let@token.\else.\null\fi\xspace}

\def\eg{\emph{e.g}\onedot} 
\def\ie{\emph{i.e}\onedot} 
 
\def\etc{\emph{etc}\onedot} \def\vs{\emph{vs}\onedot}

\makeatother

\newcommand{\otatlogo}[1]{\includegraphics[scale={#1}]{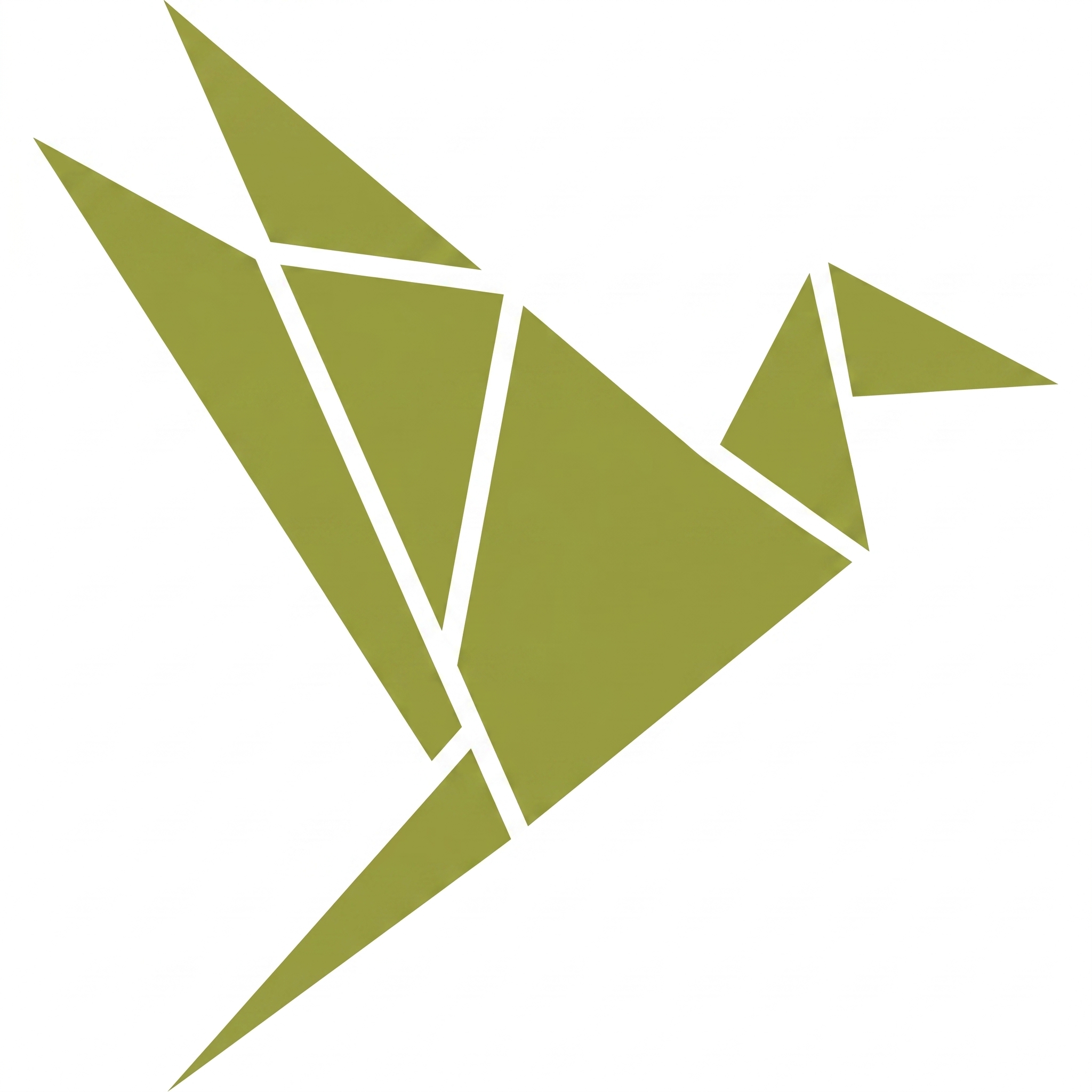}}

\title{
\otatlogo{0.009}
Attending to Multimodal Generation \\
One Token at a Time
}

\author{
Varun Gupta
\quad\quad
Vineet Gandhi
\quad\quad
Makarand Tapaswi \\
CVIT, IIIT Hyderabad \\
\vspace{-4mm}
\faChain~{\small \url{https://katha-ai.github.io/projects/otat}}
}

\begin{document}

\maketitle

\begin{abstract}
Multimodal large language models (MLLMs) generate responses autoregressively, integrating visual and linguistic information in an evolving context.
Prior work on interpretability has focused on individual layers and circuits (\emph{where}), leaving the token-level dynamics of multimodal computation \textit{during} generation (\emph{when}) underexplored.
We address this gap and study attention shifts as per semantic role;
tracking model attention to image, text, instruction, and previously generated tokens, \emph{One Token at a Time} (OTaT).
We introduce multimodal tasks that require explicit switching between visual and textual context within a single response.
Across two mainstream model families and four open-weight MLLMs of varying sizes,
we establish consistent patterns:
attention to image peaks at tokens requiring image-derived information,
instruction tokens are revisited during task transitions, and
attention to previously generated tokens increases as the generation progresses.
Causal attention blocking interventions validate the functional role of these trends.
We profile model behavior under disrupted attention and observe responses falling back to \emph{language priors}, or exhibiting \emph{cross-modal leakage}, \emph{denial}, or \emph{recovery}.
Finally, informed of the attention dynamics through our novel analysis, we propose a simple test-time intervention to boost attention to the relevant modality \emph{at the right time}, significantly improving multimodal task performance.
\end{abstract}

\section{Introduction}
\label{sec:intro}

{``\textit{A wing would be a most mystifying structure if one did not know that birds flew}'' — H. B. Barlow}

\citet{barlow1961possiblesensorymessages} noted that certain patterns become visible only when we consider how a system operates as a whole. 
This principle drives our work towards understanding Multimodal LLMs (MLLMs)~\cite{li2025llavaonevision, yang2025qwen25vl}, an essential undertaking given our increasing dependence on such systems.

The mechanistic interpretability community has made remarkable progress in examining individual attention heads, layers, and isolated circuits, revealing \textit{where} particular behaviors are implemented~\cite{elhage2021mathematicalframeworkTransCircuits, elhage2022toyModel, geva2023dissectingattentionknockout, zhang2025crossModalInfoFlow, nikankin2025sameTaskDifferentCircuits}.
However, MLLMs are \textit{dynamic} generative systems.
They do not process inputs once;
but iteratively construct outputs \textit{one token at a time}.
Each step of the forward pass uses an evolving context that includes the past outputs.

While recent efforts have begun exploring the temporal dynamics of autoregressive generation, primarily for inference efficiency~\cite{zhang2025adaptinfer, liang2025dyrate} or spatial localization~\cite{bousselham2026dexar}, comprehensive interpretability across the temporal dimension at a token-level remains underexplored.
Current works dive deep into the role of different components;
\eg~lower layers process general visual features while higher layers move towards predictions~\cite{wang2024mllmCanSee, zhang2025crossModalInfoFlow, neo2024towardsInterpreting}.
But some questions remain unanswered:
How does generation \textit{unfold} over time?
As the model produces its response, is the information routed in different ways?
Or, \emph{when} the visual and linguistic information needs to be used, what is the role of attention in \emph{orchestrating} the response?

We address these questions by studying how MLLMs generate responses.
Beyond \textit{where} models attend, we study \textul{\textit{how attention patterns evolve}} as decoding progresses.
Our approach complements finer component-level (\eg~layer) analysis.
Just as understanding \emph{flying} requires both,
the \emph{knowledge of the wing structure}, and
the \emph{principles of flight},
we posit that understanding MLLMs requires
mechanistic interpretation, \emph{and also
how computation unfolds during autoregressive generation}.

\paragraph{Summary of findings.}
We study attention patterns using a base task with minimal modality confounds, and extend our analysis to existing benchmarks, while combining visual and text processing within a single response.
We find:
(i)~Attention to \imageOrange{image} tokens peaks when generating image-relevant concepts.
Blocking attention specifically at these peaks cripples generation.
(ii)~Upon completing one task and switching to another, models revisit the \instructionRed{instruction} tokens.
Disrupting this prevents handoff (transition to the other task), resulting in partial responses.
(iii)~Attention to \previousPink{previously generated tokens} increases progressively, demonstrating that models actively leverage their own outputs to maintain fluency.
These patterns are consistent across diverse model families and generalize beyond the proposed diagnostic task to existing challenging multimodal benchmarks.
(iv)~When the task order is reversed, attention follows accordingly, hinting towards instruction-based routing.

Further, intervening on these attention patterns has interesting consequences.
\textit{Blocking} information flow from the \imageOrange{image} results in systematic patterns:
reliance on language priors,
cross-modal information leakage in smaller models,
denial of visual content,
and
an interesting behavior that enables stronger models to eventually recover from disruptions.
Conversely, \textit{boosting} attention to the right modality \textit{at the right time} results in significant performance improvements on multimodal tasks.

\paragraph{Contributions.}
(i)~We analyze token-by-token attention dynamics during autoregressive decoding.
While recent works have begun to explore this \cite{liang2025dyrate, zhang2025adaptinfer, bousselham2026dexar},
we provide the first systematic, functionally validated characterization of token semantics-aware attention orchestration during token generation in MLLMs.
(ii)~We propose a novel approach and empirical framework to compile and analyze attention dynamics for either one sample or across a dataset of multimodal tasks.
(iii)~We find that attention patterns peak at visual/linguistic context \emph{when} generating image/text-relevant answers.
We study how models transition across tasks and the importance of previously generated tokens.
All findings are validated with causal blocking interventions.
(iv)~When subjected to attention blocking, we profile how models fall back to the language prior, or exhibit cross-modal leakage or denial, or recover.
(v)~We propose a simple test-time attention boosting strategy that selectively amplifies modality-specific attention at critical decoding steps, significantly improving multimodal performance.
In summary, our work is a complementary perspective on interpreting MLLMs, which informs both theoretical understanding and practical use to improve their reliability and performance.

\section{Method}
\label{sec:method}

\begin{figure*}[t]
\centering
\small

\begin{minipage}[t]{0.45\textwidth}

A. An example from the Fruit-Math dataset.

\includegraphics[width=\linewidth]{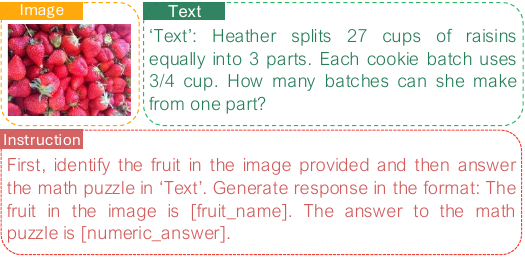}

\end{minipage}
\hfill
\begin{minipage}[t]{0.54\textwidth}

B. Information flow from input context $X$ to \cgtGreen{CGT}.

\includegraphics[width=\linewidth]{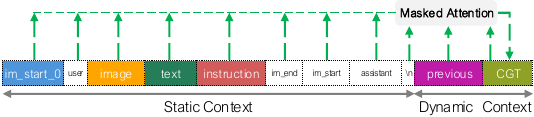}

\vspace{2mm}

C. Tagging and grouping of \textul{output tokens}.

\includegraphics[width=\linewidth]{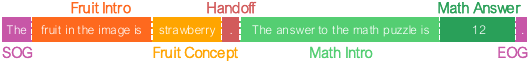}

\end{minipage}

\vspace{2mm}

\begin{minipage}[t]{0.495\textwidth}

D. Raw attention scores $\alpha_t(c_k)$ from \cgtGreen{CGT} to chunks.

\includegraphics[width=\linewidth]{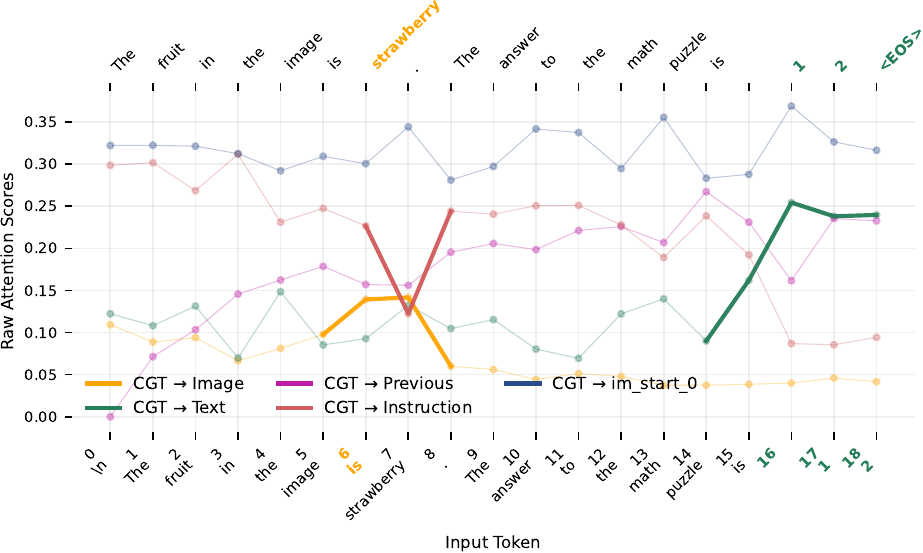}

\end{minipage}
\hfill
\begin{minipage}[t]{0.495\textwidth}

E. Normalized attention scores from \cgtGreen{CGT} to chunks.

\includegraphics[width=\linewidth]{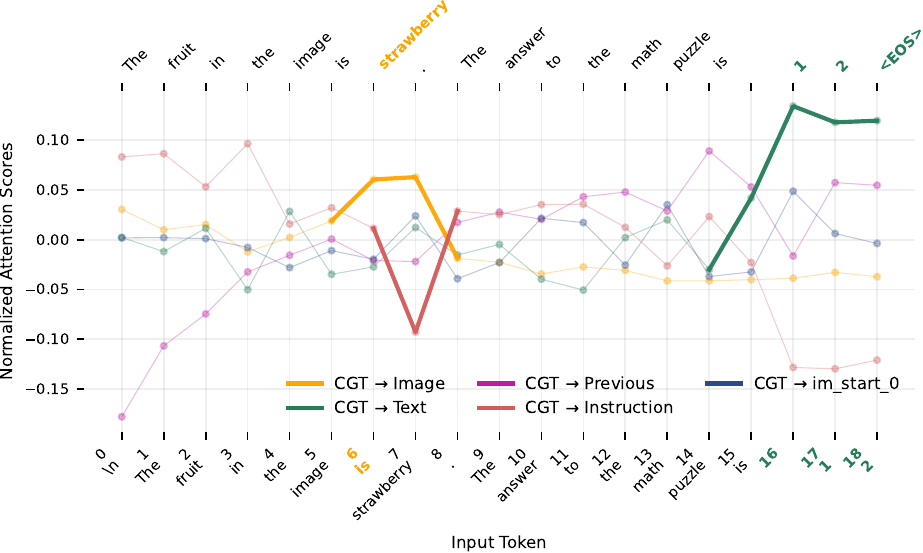}

\end{minipage}
\vspace{-1mm}
\caption{We illustrate our approach to analyze attention patterns in autoregressive MLLMs.
\textbf{A.~An example} image and text task from the Fruit-Math dataset with the response format in the instruction.
\textbf{B.~Information flow} from context to the currently generating token (\cgtGreen{CGT}).
Semantic \textit{chunks} for \lov-7B are grouped into static and dynamic context.
\textbf{C.~Tagging output} tokens and averaging attention scores within the same tag enables dataset-wide analysis and stronger conclusions.
\textbf{D.~Raw attention scores} to chunks at each token/step. 
Outputs are for \lov-7B.
Attention spikes for image tokens when answering ``strawberry'' and for text tokens when producing ``12''.
Other findings are discussed in \cref{subsec:results:attention_orchestrates}.
\textbf{E.~Attention scores normalized} by mean subtraction make it easier to interpret trends.
Best seen on screen by zooming and in color.
}
\label{fig:teaser}
\vspace{-5mm}
\end{figure*}

Starting with some background on autoregressive decoding, we propose our approach to obtain dataset-level attention patterns across multi-token responses.
We then discuss attention interventions; \textit{blocking} to establish cause-and-effect of attention trends, and
\textit{boosting} to improve MLLM performance.

\paragraph{Background.}
\label{subsec:method:background}
At a high level, a vision MLLM (\eg~\qvl~\cite{yang2025qwen25vl}) has 3 modules.
(i)~\textit{Visual encoder} converts the image into a latent representation.
(ii)~\textit{Modality projector} aligns the visual embeddings into the LLM token embedding space.
(iii)~\textit{Autoregressive decoder} (LLM) performs sequence modeling in a unified \emph{multimodal space}.
Specifically, the input context to the MLLM, consists of image embeddings for visual patches concatenated with text token embeddings.

Given input $X$, the decoder generates tokens in an autoregressive manner.
Let $y_{<t} = (y_1, \dots, y_{t-1})$ denote previously generated tokens.
At step $t$, the decoder processes the input sequence, calculates a distribution over the vocabulary $p(y_t | X, y_{<t})$, and samples $y_t$.

Let $A_t[i, j]$ refer to the attention score from the token \emph{at} index $i$ \emph{to} the token at index $j$.
MLLMs employ masked attention, \ie~$A_t [i, j] = 0$ for $j > i$.
This applies to all layers and attention heads of the decoder and ensures that each token \textit{only} attends to past context.

\subsection{Extracting Attention Patterns - One Token at a Time (OTaT) \otatlogo{0.005}}
\label{subsec:method:attention_pattern}

\paragraph{Context in MLLMs.}
It is ubiquitous in the current interpretability efforts to present a simplified view of the input context as a sequence of image and text tokens (\eg~user prompt).
However, MLLMs operate over an intricate input set, including
system instructions,
\texttt{user}/\texttt{assistant} turn indicators, 
control tokens (new line \texttt{\textbackslash n}), \etc.
For a holistic understanding, it is important to study how models attend to \emph{all} tokens and not just the modality specific inputs.
Thus, we denote the \textit{full input context} as a sequence of semantically grouped \textit{chunks}: $X = [c_1, \ldots, c_k, \ldots, c_K]$.
Note, each chunk $c_k$ may be a single token (\eg~\texttt{\textbackslash n}), tens of tokens (\eg~user instruction), or even several thousand tokens (\eg~image).
\cref{appendix:subsec:chunks} presents the list of chunks for model families studied in this work.

At decoding step $t$, the input context consists of:
(i)~\textit{Static tokens} that remain fixed through all steps
(\eg~\imageOrange{Image}, \textGreen{Text}, \instructionRed{Instruction}); and
(ii)~\textit{Dynamic tokens} that evolve during autoregressive decoding:
(a)~previously generated tokens $y_{<t}$ (\textit{\previousPink{Previous}}), and
(b)~the \textit{currently generating token} (\cgtGreen{CGT}) that will produce $y_t$ and is responsible for the attention query vector in standard QKV attention \cite{vaswani2017transformer}.

Note, during decoding, the input at \cgtGreen{CGT} is $y_{t-1}$, and the model \emph{produces} $y_t$.
Our goal is to quantify \cgtGreen{CGT}'s attention to various chunks.
Interchangeably, this also quantifies how much information flows from past tokens to \cgtGreen{CGT}.
For convenience, and without loss of generality, we extend the semantic input chunks to include the \textit{previous tokens} (\previousPink{Previous}) and \cgtGreen{CGT}.
\cref{fig:teaser}B illustrates this information flow through masked attention for \lov.

\paragraph{Attention to a chunk.}
Let $\bA_t \in \mathbb{R}^{L \times H \times N \times N}$ denote the post-softmax attention tensor across $L$ layers, $H$ heads, and between $N$ tokens at step $t$.
$N$ accounts for all tokens of the static context, $y_{1:t-1}$, and \cgtGreen{CGT} or $y_t$. 
We denote the contiguous span of tokens for chunk $c_k$ with start-end indices as $[s_{c_k}, e_{c_k})$.
\cgtGreen{CGT}'s attention to $c_k$ is:
\vspace{-1mm}
\begin{equation}
\alpha_t(c_k) =
\frac{1}{L} \sum_{l=1}^{L}
\frac{1}{H} \sum_{h=1}^{H}
\sum_{j=s_{c_k}}^{e_{c_k}-1}
A_t [l, h, t, j] \, ,
\label{eq:attnchunk}
\end{equation}
\ie~we average over layers and attention heads, and sum over all tokens in the semantic chunk. 
Although this aggregation abstracts layer and head-specific structure, the resulting signals remain informative and consistent, enabling us to study the per-token, modality-level dynamics.
These \textit{raw} attention scores for each step of the decoding process are presented in \cref{fig:teaser}D.
Note, as $\sum_{k} \alpha_t(c_k) = 1$, it provides a complete picture of how \cgtGreen{CGT} attends to all previous context.

\paragraph{Normalizing attention scores.}
The raw values of attention can be misleading~\cite{huang2024opera,kaduri2025WhatsInTheImage}.
For \eg, \cref{fig:teaser}D shows that the \imstartBlue{\texttt{im\_start\_0}} token has high attention ($\sim$0.3).
However, as the first placeholder token in the sequence, semantically, it is unlikely to provide useful information for generation 
(see blocking experiment in 
\cref{tab:blocking_metrics}).
We normalize raw attention scores by subtracting the mean attention for a chunk across all decoding steps and all samples of the dataset,
removing baseline effects and enhancing their interpretability
(\cref{fig:teaser}E).
Corresponding interpretations and results are in \cref{subsec:results:attention_orchestrates}.

\paragraph{Grouping output responses.}
\label{subsec:method:response}
Our goal is to study how attention evolves across multi-token responses.
As seen in \cref{fig:teaser}E, attention scores at each step $t$ are rich and quite insightful.
While per-sample analysis is useful, dataset-level trends enhance the robustness of findings.
This introduces a key challenge: responses vary in length and exact wording, making token-level aggregation across samples ill-posed.
We address this by mapping generated tokens $(y_1, \ldots, y_T)$ to a set of functional roles (\eg~setup, concept, handoff, \etc) defined by the task response format, illustrated in \cref{fig:teaser}C.
This aligns outputs to a shared semantic structure, enabling consistent aggregation across samples.
We then average normalized attention scores over tokens assigned the same \textit{role},
and visualize these
as bar plots to study dataset-level trends.
This preserves the semantic progression while enabling comparisons across variable-length outputs.

\begin{wrapfigure}{r}{0.3\textwidth}
\vspace{-10mm}
\centering
\includegraphics[width=\linewidth]{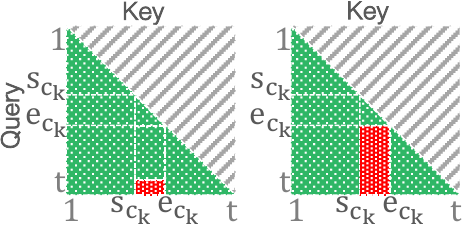}
\caption{Blocking strategies in QK attention (gray is the causal mask, red is blocked).
\textbf{Left}: Lazy blocking only affects \cgtGreen{CGT}.
\textbf{Right}: Total blocking prevents information flow to any future token.}
\vspace{-12mm}
\label{fig:method_attention_blocking}
\end{wrapfigure}

\subsection{Attention Interventions}
\label{subsec:method:totalblocking}
Attention patterns tell us the relative importance of context chunks.
However, to probe whether a chunk is \emph{functionally} important, we intervene by blocking attention to that chunk and measuring the resulting change in the model's response.
Boosting attention is the counter procedure, where we intervene to enhance the model's attention to a certain chunk or modality at a given generation step.

\subsubsection{Attention Blocking}

Blocking is straightforward when generating a single output token.
However, a few questions emerge for our multi-token output setup:

\paragraph{How to block?}
Prior works~\cite{zhang2025crossModalInfoFlow, neo2024towardsInterpreting} block attention between the current token ($y_t$) to a specific chunk (\eg~\imageOrange{image} $c_k$) at all layers and heads:
$\bA_t[:, :, t, s_{c_k}{:}e_{c_k}] = 0$.
However, information may still leak from the \imageOrange{image} to \cgtGreen{CGT}.
For example, intermediate tokens may collect image information in early layers and propagate it further to \cgtGreen{CGT} in later layers.
Thus, we refer to this as \emph{lazy blocking}.

Instead, we propose \textit{total blocking}.
This strategy prevents information flow from the chunk $c_k$ to any future token in its sequence and corresponds to setting
$\bA_t[:, :, e_{c_k}{:}t, s_{c_k}{:}e_{c_k}] = 0$ (see \cref{fig:method_attention_blocking}).

\paragraph{When to block?}
In a multi-token response, decoded tokens either primarily
follow the instruction,
ensure grammatical correctness, or
actually solve the task.
We block information flow at different steps (output tags) and analyze how models react to these interventions (see details in \cref{appendix:subsec:when_block}).

\paragraph{What to block?}
We show the impact of blocking different semantic chunks.
While previous works only study effects of blocking \emph{visual} inputs~\cite{zhang2025crossModalInfoFlow},
we analyze the importance of other chunks.

\subsubsection{Attention Boosting}
\label{subsec:attn_boosting}
If blocking attention harms generation, could amplifying it, improve outputs?
Given the post-softmax attention tensor $\bA_t$, we apply a multiplicative reweighting of attention assigned to a chunk $c_k$ at the current step $t$.
Concretely, for a boost factor $\beta$, we multiply and sum normalize attention scores:
\begin{equation}
\tilde{\bA}_t[:, :, t, j] =
\begin{cases}
\beta \cdot \bA_t[:, :, t, j], & j \in [s_{c_k}, e_{c_k}) \\
\bA_t[:, :, t, j], & \text{otherwise}
\end{cases},
\, \text{ and } \,
\bA^\beta_t[:, :, t, j] =
\frac{\tilde{\bA}_t[:, :, t, j]}
{\sum_{j'} \tilde{\bA}_t[:, :, t, j']} \, .
\label{eq:boosting}
\end{equation}

Without loss of generality, let us consider $c_k$ as the chunk that is selected for boosting.
Each token in the chunk is boosted by the same factor.
$\alpha_t(c_k)$ is the aggregated attention to that chunk (Eq.~\eqref{eq:attnchunk}).
As overall attention sums to 1, the denominator in Eq.~\eqref{eq:boosting} is: $Z_k = (1 - \alpha_t(c_k)) + \beta \cdot \alpha_t(c_k)$.
Effectively, the attention to $c_k$ is scaled by $\gamma_k = \beta / Z_k$, and $\gamma_k$ depends on the original attention mass $\alpha_t(c_k)$.
We refer to $\gamma_k$ as the effective boost factor that is experienced by each token of the selected chunk during generation of the next token.

\section{Experimental Setup}
\label{sec:expsetup}

We discuss the multimodal tasks, models, and response evaluation strategy before results and findings.

\begin{figure}[t]
\centering
\includegraphics[width=1\linewidth]{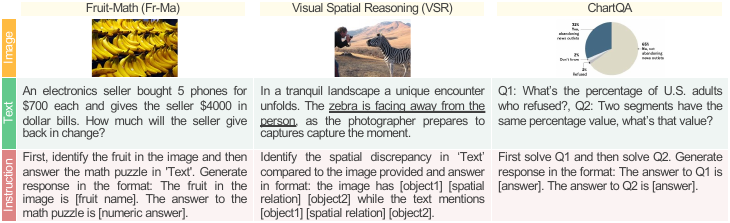}
\vspace{-3mm}
\caption{Samples from multimodal tasks.
\textbf{Fruit-Math} (left) shows an image of a fruit along with an unrelated math puzzle.
\textbf{Visual Spatial Reasoning (VSR)}~\cite{pezeshkpour2025mixedsignals} (middle) features an image with a paired caption that describes the scene, but with a conflicting spatial relationship.
\textbf{ChartQA}~\cite{masry2022chartqa} (right) shows a diagram followed by two questions related to it.
The model is instructed to (respectively):
identify the fruit in the image and solve the math puzzle,
identify the spatial discrepancy between image and text,
and answer both questions.
}
\vspace{-2mm}
\label{fig:dataset}
\end{figure}

\paragraph{Multimodal tasks.}
\label{subsec:expsetup:dataset}
To rigorously assess the dynamics of generation, we analyze models on a suite of multimodal tasks.
\cref{fig:dataset} shows one sample for each task and \cref{appendix:subsec:dataset} presents more examples.

\begin{enumerate}[noitemsep,topsep=0pt,leftmargin=*]
\item \textbf{Fruit-Math} (Fr-Ma)
is used as a diagnostic task and has samples across 11 fruit categories from the OpenImages dataset~\cite{OpenImages}, paired (randomly) with math puzzles from GSM-8K~\cite{cobbe2021gsm8k}.
The model is tasked to identify the fruit in the image and solve the math puzzle.
By enforcing non-overlapping information in modalities, we eliminate redundancy as a confounder to trace the contribution of each input chunk independently.

\item \textbf{Visual Spatial Reasoning} (VSR)
is adapted directly from the Mixed Signals benchmark~\cite{pezeshkpour2025mixedsignals}.
Here, the image and paired text caption convey diverging spatial relations between key objects in the image (\eg~zebra facing \underline{towards} \vs~\underline{away} from the photographer).
Given the conflicting sample, the model is instructed to identify the spatial relation in the image and in the text.

\item \textbf{ChartQA}~\cite{masry2022chartqa} is a traditional VQA benchmark that requires reasoning about data visualizations.
The benchmark contains two questions for each chart.
We provide the model with the chart image and corresponding questions, and prompt it to solve both questions as a single response.
\end{enumerate}

For all tasks, we adopt a response format to perform dataset-level aggregation and analysis.
Please note, this is \textit{not} required to analyze per-sample attention patterns, which are independently insightful.

\paragraph{Models.}
\label{subsec:expsetup:models}
We analyze outputs from two families of state-of-the-art MLLMs:
\lov{} (\lovshort)~\cite{li2025llavaonevision} and
\qvl{} (\qvlshort)~\cite{yang2025qwen25vl}.
Specifically, 0.5B and 7B for \lovshort{}, and 3B and 7B for \qvlshort{}.
Thus, our experiments feature models with differing base LLMs, \#parameters, order of input chunks, training strategies, data, and methods, and we successfully identify common attention trends across them.
We limit generation to $25$ tokens for Fr-Ma, $128$ for VSR and ChartQA, or until $\langle\texttt{EOS}\rangle$.

\paragraph{Compiling attention scores across responses.}
\label{subsec:expsetup:responses}
We tag and group output tokens for dataset-level aggregation and reporting.
An example grouping for Fr-Ma is illustrated in \cref{fig:teaser}C.
We average attention scores of tokens belonging to the same tag to report robust observations across the entire dataset.
The Gemini prompts used to tag outputs are in \cref{appendix:subsec:gemini_pos_prompts}.
To prevent clutter here, the specific output token groups for Fr-Ma, VSR, and ChartQA are presented in \cref{appendix:subsec:pos_tags}.

To perform quantitative evaluation of model responses, we evaluate answers using Gemini 2.5 Pro
(prompts in \cref{appendix:subsec:gemini_eval_prompts}).
We perform spot-checks to ensure high reliability of this automatic approach.
\section{Results and Findings}
\label{sec:results}

We present common trends observed across models and tasks, followed by causal effects of total attention blocking.
Finally, we demonstrate how attention boosting improves model performance.

\begin{figure}[t]
\includegraphics[width=0.49\linewidth]{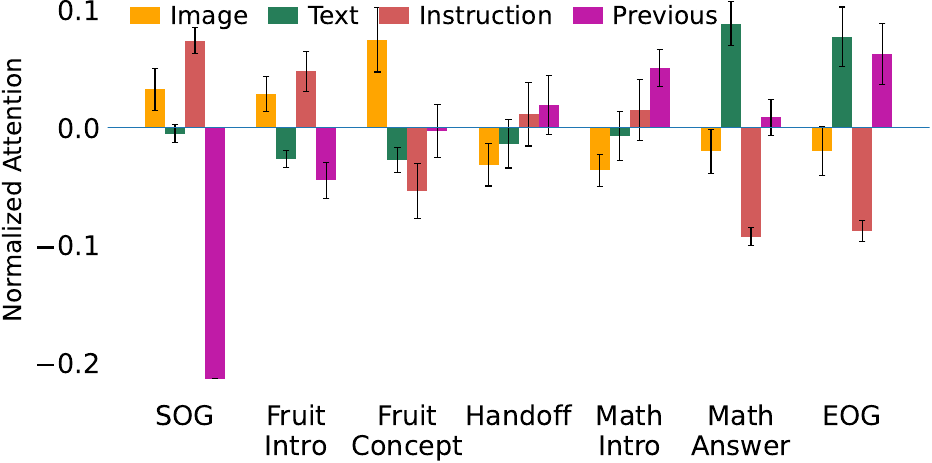}
\hfill
\includegraphics[width=0.49\linewidth]{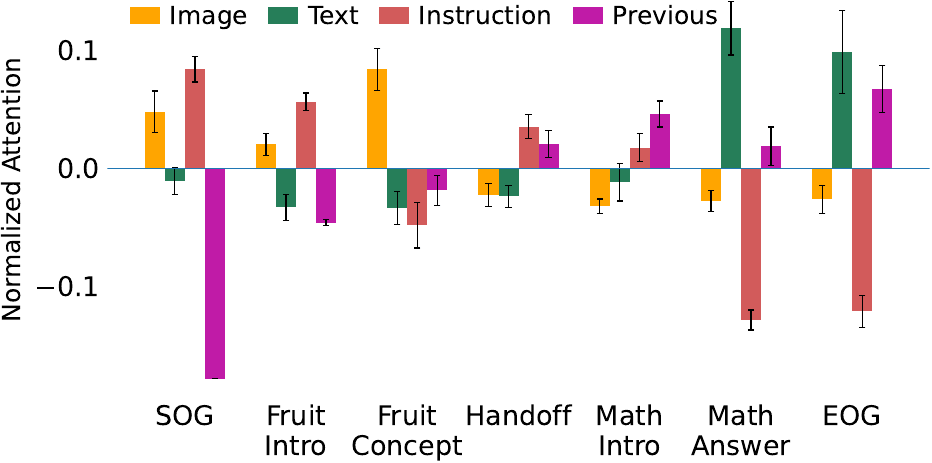}
\vspace{-1mm}
\caption{Evolution of attention patterns in Fr-Ma for \qvlshort-3B (left) and \lovshort-7B (right).}
\label{fig:attention_fruit-math_main}
\vspace{-3mm}
\end{figure}

\subsection{Attention Orchestrates Autoregression}
\label{subsec:results:attention_orchestrates}

We visualize per-token attention scores for a given data sample in \cref{fig:teaser}D, and the normalized scores in \cref{fig:teaser}E.
Additional examples are in \cref{appendix:subsec:line_plots},
along with a grouped layer-level analysis that validates our global averaging approach in \cref{appendix: subsec: layer_grouped_analysis}.
While sample-level analysis (line plots) provides similar insights, we extend observations to the entire dataset (bar plots) by tagging output tokens and averaging attention scores within tags.

We discuss attention trends across \imageOrange{image}, \textGreen{text}, \instructionRed{instruction}, and \previousPink{previous} chunks, and identify common patterns that hold across models/tasks.

\paragraph{Models attend to modality-relevant context when required.}
\cref{fig:attention_fruit-math_main} presents attention patterns for \qvlshort-3B and \lovshort-7B on Fr-Ma.
Attention peaks to the image chunk when producing the fruit name, \ie~\textit{Fruit Concept},
is low-to-moderate when generating \textit{Fruit Intro}:  (``the fruit in the image is''), and
falls off when producing \textit{Math Answer}.
A similar trend is observed in text.
Attention to the text chunk (math puzzle) peaks when producing \textit{Math Answer} and is low elsewhere.

\begin{inlinebox}
\centering
\ding{192}
MLLMs dynamically reallocate attention to the relevant modality during generation.
\end{inlinebox}

\begin{figure}[b]
\centering
\begin{minipage}{0.49\textwidth}
\vspace{-2mm}
\includegraphics[width=\linewidth]{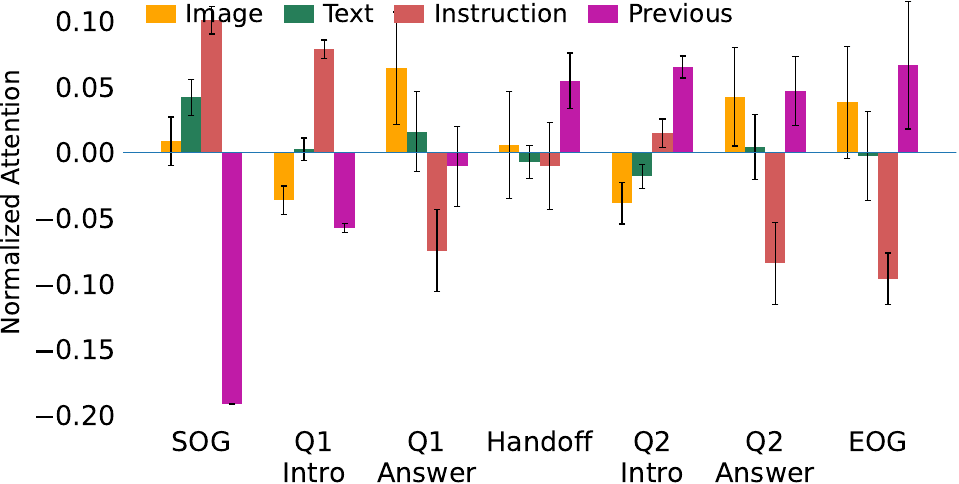}
\caption{Attention trends for ChartQA, \lovshort-7B.
Image is attended when generating Q1 and Q2 answer.}
\label{fig:attention_chartqa_main}
\end{minipage}
\hfill
\begin{minipage}{0.49\textwidth}
\vspace{-2mm}
\includegraphics[width=\linewidth]{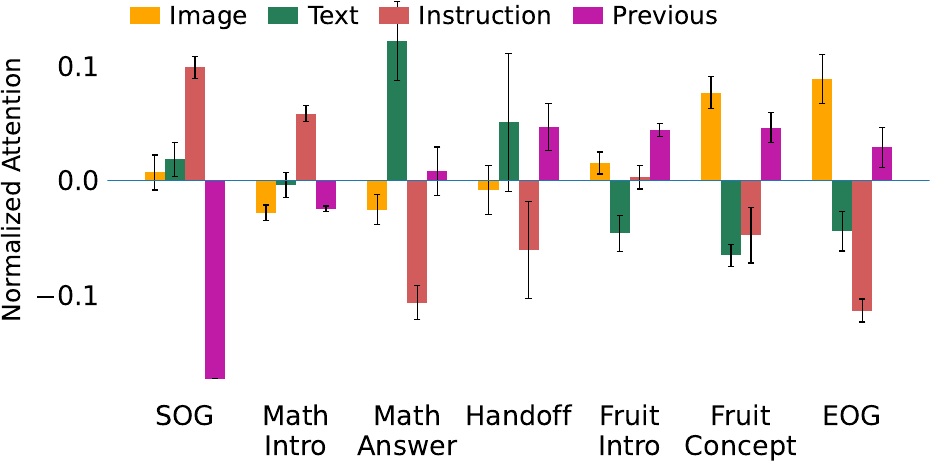}
\caption{Attention trends flip when the response format is reversed (``Math-Fruit'' task for \lovshort-7B).}
\label{fig:attention_mathfruit}
\end{minipage}
\end{figure}

\paragraph{Reversing task order flips attention trends.}
To confirm that the above patterns are not a spurious artifact,
for the same input context, we instruct the model to first solve the math puzzle and then identify the fruit.
Comparing \cref{fig:attention_fruit-math_main}(bottom) \vs~\cref{fig:attention_mathfruit},
we see that attention patterns mirror this reversal.
Models attend to \textGreen{text} for \textit{Math Answer} and \emph{then} to \imageOrange{image} for \textit{Fruit Concept}.

\paragraph{Task switching revisits instruction.}
There are two spikes to the instruction chunk: once at the start (\textit{SOG}) and later during the change in tasks from identifying the fruit in the image to solving the math puzzle, \ie~at or right after the \textit{Handoff}.
This explains how models generate multi-answer responses.

\begin{inlinebox}
\ding{193}
MLLMs revisit instruction tokens when switching tasks (\eg~visual perception to arithmetic).
\end{inlinebox}

\paragraph{Attention to previously generated tokens}
increases consistently as decoding progresses, hinting that previous tokens assist in maintaining grammatical structure and adherence to format.

All above trends are also observed for other models (\cref{appendix:subsec:attention_trends}).
In fact, base LLMs exhibit such attention patterns too, given two language contexts and tasks
(details in \cref{appendix:subsec:llm_only_experiment}).

\paragraph{When does attention peak?}
In addition to attention peaking at the right moments, we see attention for text spike at \textit{EOG} in \cref{fig:attention_fruit-math_main}.
A similar spike is seen for the image chunk at \textit{EOG} in \cref{fig:attention_mathfruit}.
As explained earlier, the \cgtGreen{CGT} produces $y_t$ with input $y_{t-1}$.
Feeding back the correctly decoded fruit ($y_{t-1}$) triggers the attention scoring mechanism, as the concept \textit{is} related to the \imageOrange{image}.
Thus, it increases the attention score for CGT even while producing \texttt{$y_t=\langle$EOS$\rangle$}.
This phenomenon is more clearly observed in the per-token plots (\cref{fig:teaser}D).
Attention to the \imageOrange{image} chunk is high while decoding ``strawberry'', but also when ``strawberry'' is fed back as input for the next step.

\begin{inlinebox}
\centering
\ding{194}
Attention scores are influenced by what the model is decoding ($y_t$) \textit{and} the input ($y_{t-1}$).
\end{inlinebox}

\paragraph{ChartQA.}
We present results on a traditional VQA task, ChartQA, modified to answer both questions.
Unlike Fr-Ma, ChartQA is \textit{entangled}, with the text chunk asking questions relevant to the image.
\cref{fig:attention_chartqa_main} confirms the above findings:
The model attends to the \imageOrange{image} twice, reflecting chart glimpses when answering the two questions.
Attention to \previousPink{previous} tokens consistently increase, and around the \textit{Handoff} token, attention to \instructionRed{instruction} tokens also spike.
This suggests that the previous findings are not specific to disentangled modality tasks (Fr-Ma), and also persist for ChartQA.

\paragraph{Consistent and high attention scores.}
Certain model variants (\lovshort-7B, \qvlshort-3B) show high attention scores for the first token in the input context, \imstartBlue{\texttt{im\_start\_0}} (blue line in \cref{fig:teaser}D).
As the first context token does not contain meaningful information, the high score is unexplained.
However, post normalization (\cref{fig:teaser}E), the score is close to 0, indicating that this has little impact on generation.

\begin{inlinebox}
\ding{195}
Raw attention scores may be high for possibly unimportant tokens.
However, via normalization, studying whether attention scores change across timesteps is helpful to gain insights.
\end{inlinebox}

\subsection{Effects of Blocking Information Flow}
\label{subsec:results:attention_blocking}

We analyze causal relationships of attention trends observed in Fr-Ma by blocking.

\begin{wraptable}{R!}{0.67\textwidth}
\small
\centering
\tabcolsep=0.06cm
\vspace{-4mm}
\caption{Fruit accuracy across all models is impacted by attention blocking when decoding \textit{Fruit Concept} (FC), but unaffected by blocking during \textit{Fruit Intro} (FI).
Reporting mean$\pm$stddev and change from no blocking.}
\vspace{-2mm}
\label{tab:attention_blocking_image}
\begin{tabular}{cc llll}
\toprule
\multicolumn{2}{c}{Blocking} & \multicolumn{2}{c}{\cellcolor{yellow!5}\lovshort} & \multicolumn{2}{c}{\cellcolor{orange!5}\qvlshort} \\
\multicolumn{1}{c}{How?} &
\multicolumn{1}{c}{When?} &
\multicolumn{1}{c}{0.5B} &
\multicolumn{1}{c}{7B} &
\multicolumn{1}{c}{3B} &
\multicolumn{1}{c}{7B} \\
\midrule
\multicolumn{2}{c}{No Blocking}
  & 65.1 {\miniscule $\pm$1.4 }
  & 64.9 {\miniscule $\pm$1.5 }
  & 64.8 {\miniscule $\pm$1.5 }
  & 65.3 {\miniscule $\pm$1.3 } \\
Lazy & FC
  & \phantom{0}6.4 \reducenum{58.7} {\miniscule $\pm$0.7 }
  & 33.4 \reducenum{31.5} {\miniscule $\pm$1.4 }
  & 36.1 \reducenum{28.7} {\miniscule $\pm$1.3 }
  & 46.7 \reducenum{18.6} {\miniscule $\pm$1.5 } \\
Total & FC
  & 31.1 \reducenum{34.0} {\miniscule $\pm$1.5 }
  & 15.0 \reducenum{49.9} {\miniscule $\pm$1.1 }
  & 37.3 \reducenum{27.4} {\miniscule $\pm$1.3 }
  & 28.5 \reducenum{36.8} {\miniscule $\pm$1.3 } \\
Total & FI
  & 69.1 \increasenum{+3.9} {\miniscule $\pm$1.1 }
  & 70.9 \increasenum{+6.1} {\miniscule $\pm$1.1 }
  & 65.5 \similarnum{+0.7} {\miniscule $\pm$1.2 }
  & 69.5 \increasenum{+4.2} {\miniscule $\pm$1.1 } \\
\bottomrule
\end{tabular}
\vspace{-4mm}
\end{wraptable}

\paragraph{On fruit identification.}
\cref{tab:attention_blocking_image} (row 1) presents fruit accuracy of all 4 models \emph{without} blocking.
Next, we block the information flow when \cgtGreen{CGT} is decoding the \textit{Fruit Concept} via lazy or total blocking from the image chunk to \cgtGreen{CGT}.
All models see a large drop in accuracy (rows 2, 3), while larger models (7B) are affected more by total blocking than lazy blocking.
In row 4, we observe that blocking attention to the image when generating tokens of \textit{Fruit Intro}
leaves the fruit accuracy mostly unaffected (small improvements).
The blocking experiment confirms Finding \ding{192}, that attention towards the image chunk is necessary, but primarily \textit{when} decoding the fruit name.

\begin{wrapfigure}{R}{0.7\textwidth}
\centering
\includegraphics[width=0.49\linewidth]{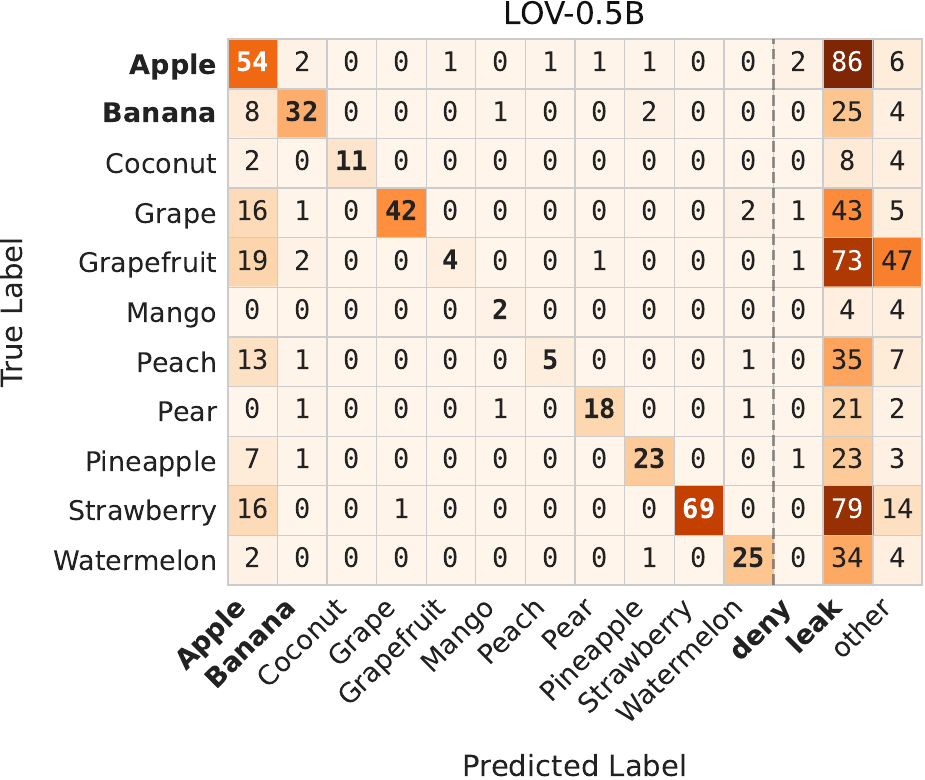}\hfill
\includegraphics[width=0.49\linewidth]{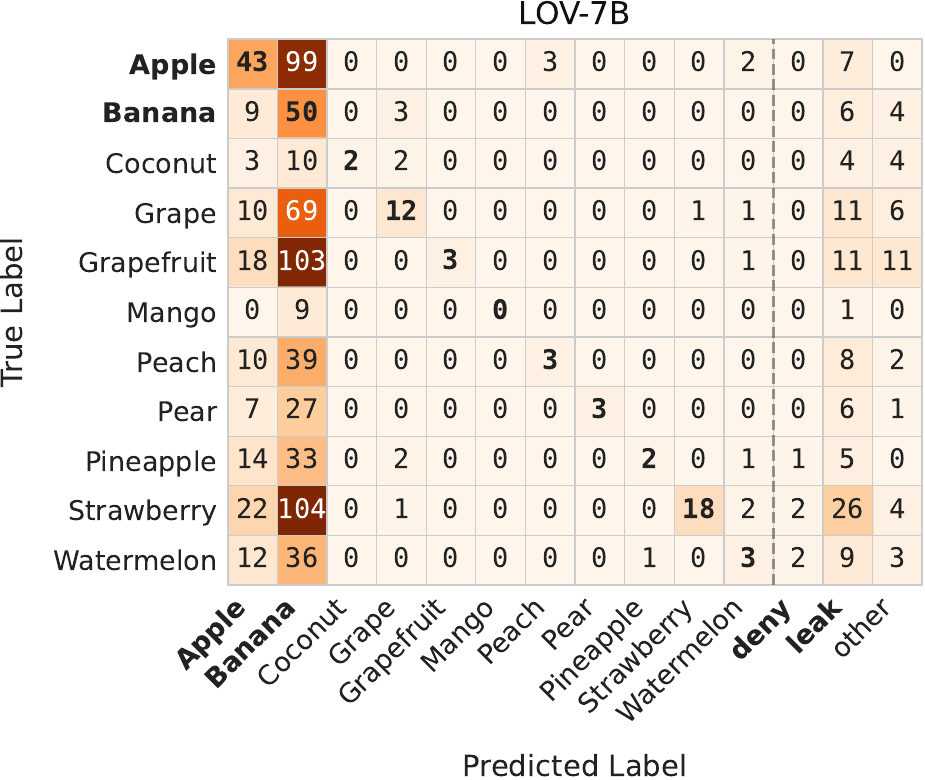} \\
\includegraphics[width=0.49\linewidth]{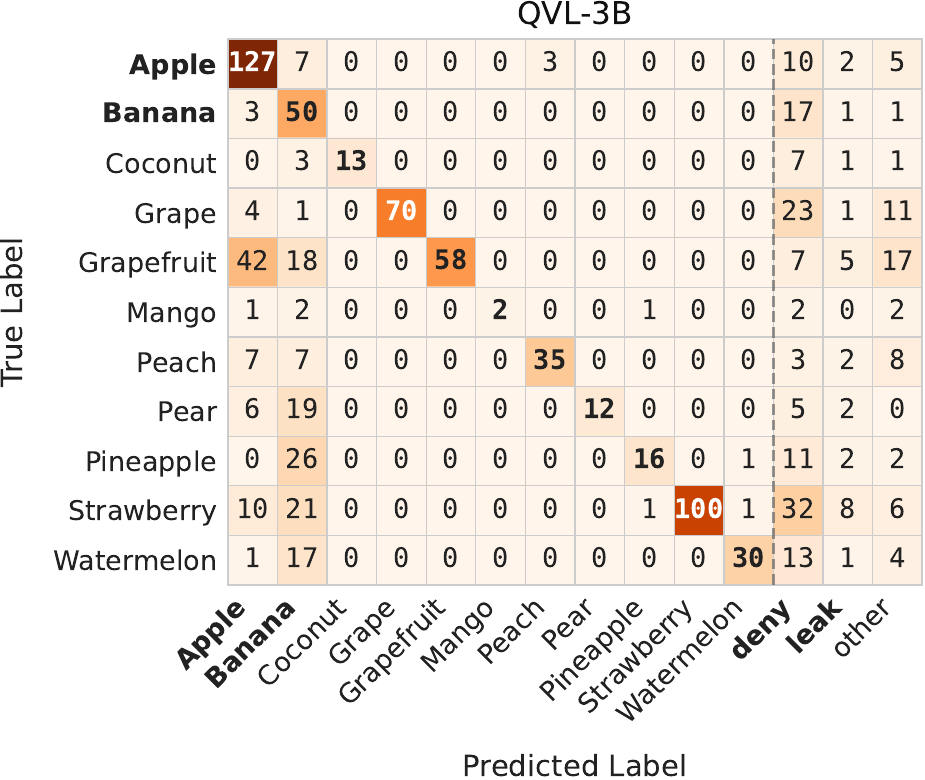}\hfill
\includegraphics[width=0.49\linewidth]{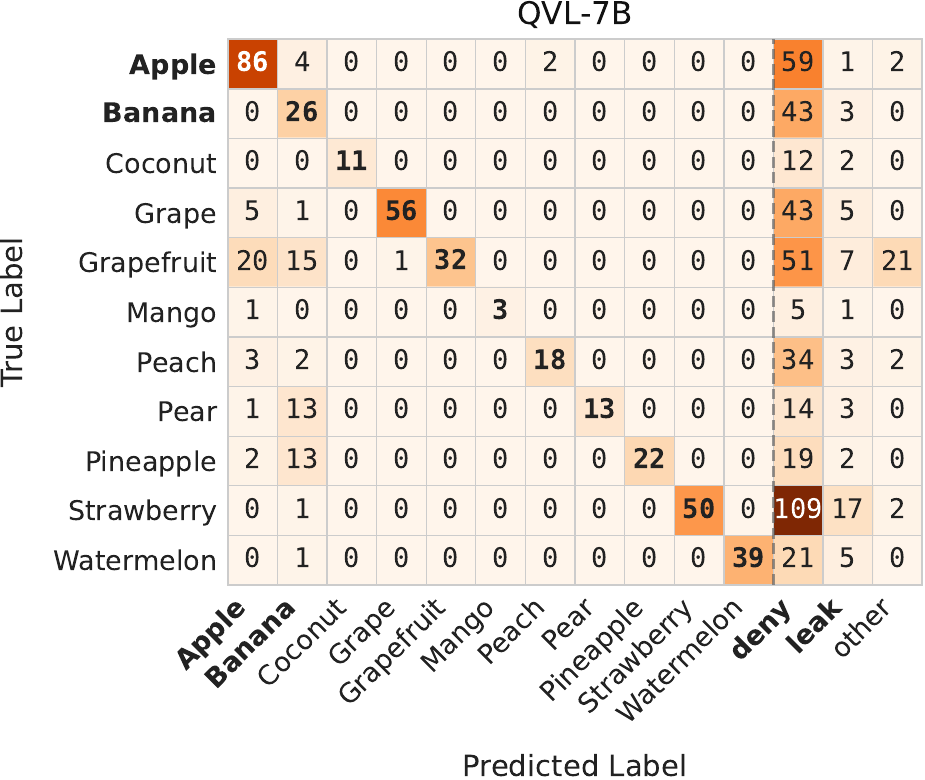}\hfill
\vspace{-1mm}
\caption{
Confusion matrices showing models' outputs after blocking at \textit{Fruit Concept}.
\lovshort-0.5B shows leakage of cross-modal information (top-left).
\lovshort-7B falls back to the language prior (top-right).
\qvlshort-3B exhibits mixed signs of language prior, denial, and recovery (bottom-left).
\qvlshort-7B exhibits strong denial and recovery (bottom-right).
Best on screen.
}
\label{fig:blocking_confusion}
\vspace{-2mm}
\end{wrapfigure}

\paragraph{Profiling outputs of blocking at \textit{Fruit Concept}.}
With total blocking, the model is completely unaware of image information at that timestep.
So why is fruit accuracy not 0?
Additionally, in \cref{tab:attention_blocking_image}, why does \lovshort-0.5B show higher accuracy for total \vs~lazy blocking?
We analyze model outputs through confusion matrices (see \cref{fig:blocking_confusion}), and identify 4 typical response categories:

\begin{enumerate}[noitemsep, topsep=0pt, leftmargin=*]
\item Small models \textbf{leak} nouns present in the math puzzle (\eg~money) and generate bizarre responses such as ``fruit in the image is \textit{money}''.
For \lovshort-0.5B, 50-60\% of responses contain leaked terms.

\item Using \textbf{language priors} results in some correct answers.
\lovshort-7B shows a particularly high affinity for popular fruit \textit{apple}, and \textit{banana}.

\item Models \textbf{deny} existence of a fruit with responses such as ``fruit in the image is \textit{not applicable} [...]''.
Compared to \lovshort, \qvlshort-7B, exhibits a high denial rate and may have been trained for this.

\item \qvlshort{} models are able to \textbf{recover} after blocking at a specific timestep with responses such as ``fruit in the image is \textit{not oranges}, its a close-up of oranges'', which is considered correct by the scorer.
\end{enumerate}

This explains the non-zero performance of most models.
Specific to \lovshort-0.5B, lazy blocking results in higher denial and leakage as compared to total blocking, reducing accuracy, as such outputs are always marked incorrect.
Qualitative examples across categories are presented in \cref{appendix:subsec:blocking_qualitative_fruit}.

\begin{inlinebox}
\ding{196}
MLLMs fall back to language \emph{priors}, \emph{leak} cross-modal information, \emph{deny} visual content, or sometimes, exhibit \emph{recovery} behaviors when attention to the image is blocked at the critical stage.
\end{inlinebox}

\phantom{0}

\paragraph{Blocking beyond images.}
\cref{tab:attention_blocking_other} presents results for blocking different context chunks (\texttt{what}) at different decoding steps (\texttt{when}).
Beyond accuracy, we also quantify whether the model generates a valid answer.
(i)~Comparing rows 1 and 2, we see that blocking information flow from \imstartBlue{\texttt{im\_start\_0}} does not decrease performance, confirming Finding \ding{195}.
In fact,
the model's performance \textit{improves slightly} on eliminating attention to \textit{passive} chunks.
(ii)~Next, row 3 shows the effect of blocking attention to the \instructionRed{instruction} chunk when \cgtGreen{CGT} is at the \textit{Handoff}.
While fruit performance is unaffected, models stop answering the math question (0.2\% and 1.2\%).
This confirms Finding \ding{193} that models revisit the instruction chunk for task switching.
(iii)~Finally, row 4 shows that blocking attention to \previousPink{previous}
breaks responses.
Models get stuck (\eg~``\textit{the fruit the fruit ...}''), and the fruit answering rate plummets.
While \lovshort-7B is more sensitive (25.9\% rate, 67.5\% accuracy),
\qvlshort-3B is more robust to blocking and performs reasonably (67.5\% rate, 47.4\% accuracy).
However, both models fail to produce the math answer.
Qualitative examples are in \cref{appendix:subsec:blocking_other}.
\begin{table}[t]
\small
\centering
\tabcolsep=0.5mm
\caption{Effect of total blocking of semantic chunks (\textit{what}) at different decoding steps (\textit{when}) on the Fruit-Math task.
``F/M-Acc'' is fruit/math accuracy; and
``F/M-Ans?'' is the answer rate of the fruit/math question.
We report results for two models:
M$_1$ is \lovshort-7B and M$_2$ is \qvlshort-3B.}
\label{tab:attention_blocking_other}
\label{tab:blocking_metrics}
\begin{tabular}{cc @{\hspace{3mm}}  ll @{\hspace{3mm}} ll @{\hspace{3mm}} ll @{\hspace{3mm}} ll}
\toprule
\multicolumn{2}{c}{Blocking} & \multicolumn{2}{c}{F-Ans?} & \multicolumn{2}{c}{F-Acc} & \multicolumn{2}{c}{M-Ans?} & \multicolumn{2}{c}{M-Acc} \\
\multicolumn{1}{c}{What} & \multicolumn{1}{c}{When} &
\multicolumn{1}{c}{M1} & \multicolumn{1}{c}{M2} &
\multicolumn{1}{c}{M1} & \multicolumn{1}{c}{M2} &
\multicolumn{1}{c}{M1} & \multicolumn{1}{c}{M2} &
\multicolumn{1}{c}{M1} & \multicolumn{1}{c}{M2} \\
\midrule
\multicolumn{2}{c}{No Blocking}
  & \phantom{0}93.8 & 94.0
  & 64.9 & 64.8
  & \phantom{0}83.9 & 84.2
  & \phantom{0}3.8 & \phantom{0}3.7 \\
\imstartBlue{\texttt{im\_start\_0}} & All
  & 100 \increasenum{6.2} & 99.9 \increasenum{5.9}
  & 71.1 \increasenum{6.2} & 67.0 \increasenum{2.2}
  & 100 \increasenum{16.1} & 91.3 \increasenum{7.1}
  & 13.6 \increasenum{9.8} & 12.2 \increasenum{8.5} \\
\instructionRed{Instruction} & Handoff
  & 100 \increasenum{6.2} & 99.8 \increasenum{5.8}
  & 70.7 \increasenum{5.8} & 66.3 \increasenum{1.5}
  & \phantom{00}0.2 \reducenum{83.7} & \phantom{0}1.2 \reducenum{83.0}
  & \phantom{0}0.0 \reducenum{3.8} & \phantom{0}0.1 \reducenum{3.6} \\
\previousPink{Previous} & All
  & \phantom{0}25.9 \reducenum{67.9} & 67.5 \reducenum{26.5}
  & 16.3 \reducenum{48.6} & 47.4 \reducenum{17.4}
  & \phantom{00}2.7 \reducenum{81.2} & \phantom{0}0.7 \reducenum{83.5}
  & \phantom{0}0.0 \reducenum{3.8} & \phantom{0}0.0 \reducenum{3.7} \\
\bottomrule
\end{tabular}
\vspace{-4mm}
\end{table}

\subsection{Application: Token-Level Attention Guidance Improves Performance}
\label{subsec:results:attention_boosting}
\begin{figure}
\centering
\includegraphics[width=0.49\linewidth]{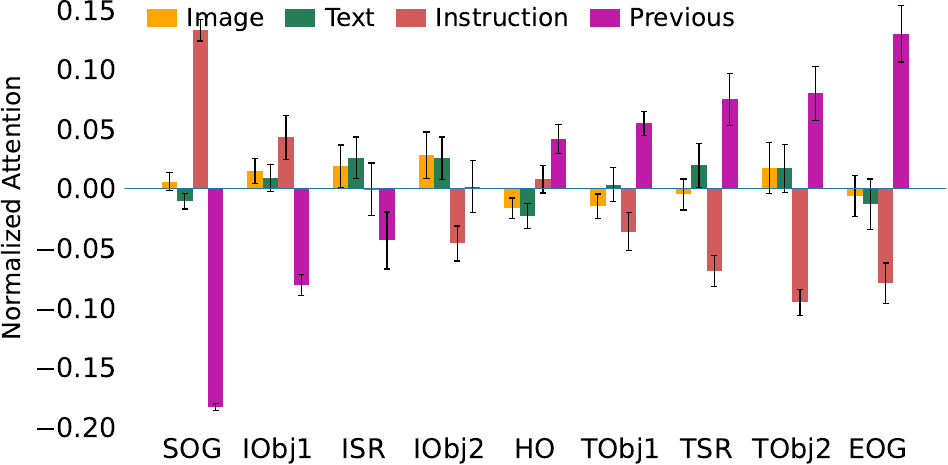}\hfill
\includegraphics[width=0.49\linewidth]{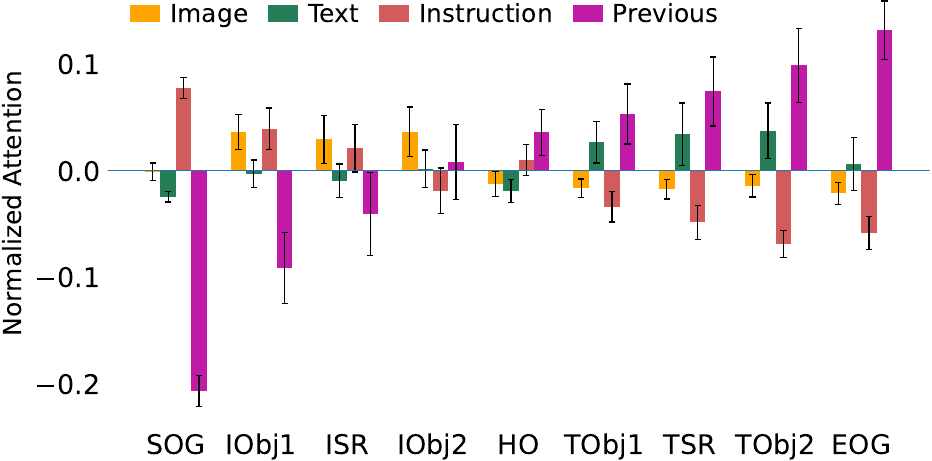}
\caption{Attention trends for the VSR task explain why \qvlshort-7B (right) outperforms \lovshort-7B (left), with peaks to image and text modalities at the appropriate image/text spatial relation (ISR, TSR) tags.}
\vspace{-4mm}
\label{fig:attention_vsr}
\end{figure}

VSR is harder than Fr-Ma due to the entangled nature of the image and text modalities and fine differences.
We observe that \lovshort-7B performs worse than \qvlshort-7B (\cref{tab:attn_boost} 21.3\% \vs~50.7\%).
Notably, our attention trends analysis explains this disparity.
In \cref{fig:attention_vsr}, we see that \lovshort{} attends predominantly to \textGreen{text} rather than \imageOrange{image} when generating both, image and text spatial relationship (ISR, TSR).
In contrast, \qvlshort~exhibits the expected pattern with attention to \imageOrange{image} peaking at ISR and \textGreen{text} at TSR.
Motivated by this, we ask: Would performance improve if we amplified attention to the correct modality at the right time?

\paragraph{\textit{When} to boost?}
Our attention analysis framework provides clear guidance on when models should attend to a specific modality.
As an intervention, we propose to boost attention to the image or text at the first ISR or TSR token, respectively.
We also consider a baseline that boosts attention to the image chunk (neglected modality) throughout generation.
We apply boosting naively across all layers and heads for all interventions.
\cref{tab:attn_boost} shows that always boosting attention to the image chunk (Img) largely degrades performance across all models, especially resulting in large drops in text accuracy (\lovshort-7B -41.2\%, \qvlshort-3B -33.9\%, \qvlshort-7B -50.7\%).
Even image accuracies witness significant drops (\lovshort-7B -9.2\%, \qvlshort-7B -26.7\%).
Surprisingly, the accuracy increases by 0.5\% for the \qvlshort-3B model, possibly indicating a larger attention deficit on the image modality, which increases when globally boosted. 
Thus, role-agnostic intervention disrupts task performance.

\begin{wraptable}{T}{0.48\textwidth}
\vspace{-4mm}
\small
\centering
\tabcolsep=0.2cm
\caption{VSR results of attention boosting.
V-/T-/VT-Acc are vision, text, both accuracy.
}
\label{tab:attn_boost}
\vspace{-2mm}
\begin{tabular}{lc c c c}
\toprule
\multicolumn{1}{c}{M} &
\multicolumn{1}{c}{Boost$_{\scriptstyle \gamma}$} &
\multicolumn{1}{c}{V-Acc} &
\multicolumn{1}{c}{T-Acc} &
\multicolumn{1}{c}{VT-Acc} \\
\midrule
\multirow{3}{*}{
\raisebox{0pt}[\height][\depth]{\rotatebox{90}{\scriptsize \lovshort-7B}}}
& - & 24.3 & 54.1 & 21.3  \\
& Img$_{\scriptstyle 11}$
& 15.1 \reducenum{9.2}  & 12.9 \reducenum{41.2} & 8.6 \reducenum{12.7} \\
& Ours$_{\scriptstyle 11}$
& 52.9 \increasenum{28.6} & 84.5 \increasenum{30.4} & 49.8 \increasenum{28.5} \\
\midrule

\multirow{3}{*}{
\raisebox{0mm}[\height][\depth]{\rotatebox{90}{\scriptsize \qvlshort-3B}}}
& - & 31.3 & 63.9 & 28.5 \\
& Img$_{\scriptstyle 13}$
& 31.8 \increasenum{0.5} & 30.0 \reducenum{33.9} & 16.5 \reducenum{12} \\
& Ours$_{\scriptstyle 13}$ & 39.6 \increasenum{8.3}  & 84.5 \increasenum{20.6}  & 37.1 \increasenum{8.6} \\
\midrule

\multirow{3}{*}{
\raisebox{0pt}[\height][\depth]{\rotatebox{90}{\scriptsize \qvlshort-7B}}}
& - & 55.7 & 81.4 & 50.7 \\
& Img$_{\scriptstyle 13}$
& 29.0 \reducenum{26.7} & 30.7 \reducenum{50.7} & 16.1 \reducenum{34.6} \\
& Ours$_{\scriptstyle 13}$
& 59.3 \increasenum{3.6}  & 87.6 \increasenum{6.2}  & 53.4 \increasenum{2.7} \\
\bottomrule
\end{tabular}
\vspace{-4mm}
\end{wraptable}

In contrast, boosting the right modality at the right time (Ours) results in the best improvements for vision, text, and both (\lovshort-7B +28.5\%, \qvlshort-3B +8.6\%, \qvlshort-7B +2.7\%) accuracy.
Overall, this result shows that attention should be modulated in a timestep aware manner; indiscriminate boosting is harmful, and targeted interventions are effective even without factoring in architecture specifics (such as layers and heads), which are beyond the scope of this work.

\paragraph{\textit{How much} to boost?}
We sweep over the effective boost factors and empirically observe optimal $\gamma {\sim=} 11$ for \lovshort{} and $\gamma {\sim=} 13$ for \qvlshort{} (\cref{tab:attn_boost}).
We observed that QVL-7B has lower attention to the image chunk as compared to \lovshort-7B;
possibly explaining the need for a higher $\gamma$. 
Detailed sensitivity analysis of $\beta$ and an explanation of how $\gamma$ is computed are in \cref{appendix:subsec:boost_factor_gamma}.
Across a range of values,
VT-Acc varies by less than 1.1\% on both models, indicating that the method is \textit{not} sensitive to the boost factor.
However, performance starts to gradually degrade at very high values for $\beta$.

\section{Related Work}
\label{sec:related_work}
We present an overview of existing efforts on 
multimodal interpretability,
attention-based analysis,
autoregressive dynamics,
and attention interventions to enhance performance.

\paragraph{Multimodal interpretability.}
Recent works probe the internal mechanisms of MLLMs.
Existing literature suggests that visual signals undergo a redistribution across layers, though there is limited consensus about the localization of visual information, its alignment with language, and its utilization.
\citet{neo2024towardsInterpreting} imply progressive alignment of vision with language, while \cite{yin2025liftingtheveil, nikankin2025sameTaskDifferentCircuits,wang2024mllmCanSee} indicate visual grounding occurs early and is transformed, or even attenuated, in the later layers.
Some works identify intermediate layers to be central for cross-modal interaction, where task-relevant information is consolidated before being propagated~\cite{kaduri2025WhatsInTheImage,zhang2025crossModalInfoFlow}.
While these studies provide valuable insights into \textit{where} and \textit{how} modalities interact \textit{within} the model, they focus on component-specific, layer or head-wise, static analyses.
In contrast, we do not isolate individual layers or heads, and adopt a global perspective to study how multimodal context is dynamically leveraged during autoregressive generation, which remains underexplored.

Interpreting multimodal models remains challenging due to representation entanglement across modalities~\cite{huh2024platonicrepresentation, luo2024vlmcreatecrossmodaltask, hua2025VLMConfictingInfoModal}.
To mitigate such confounds, prior work has introduced controlled benchmarks ~\cite{wenderoth2025intershap, nikankin2025sameTaskDifferentCircuits}.
In the same spirit, we design Fr-Ma such that modalities contribute independently.
Note that our findings also extend to real-world multimodal tasks: ChartQA~\cite{masry2022chartqa} and VSR~\cite{pezeshkpour2025mixedsignals}.

Prior works show that attention \cite{vaswani2017transformer, dosovitskiy2020image} supports meaningful explanations when paired with appropriate experimental controls and causal interventions \cite{clark2019whatdoesbertlookat, geva2023dissectingattentionknockout, neo2024towardsInterpreting, zhang2025crossModalInfoFlow, yin2025liftingtheveil, kaduri2025WhatsInTheImage}.
We adopt this causal perspective; we not only \emph{observe} interpretable attention patterns, but \textit{validate} their functional role by studying how blocking them disrupts the behaviors they support.

\paragraph{Autoregressive dynamics.}
Limited work examines autoregressive behavior in language and multimodal models.
\citet{oh2023tokenllmautoregression} analyze autoregression in LLMs via linear decompositions and changes in token probabilities, and characterize models as largely correlational, lookup-driven predictors.
In the multimodal setting, EAGLE~\cite{chen2025mllmsWhereMLLMAttend} 
attributes outputs to visual or textual evidence using a black-box, post-hoc framework.

Recent works explore the internal attention dynamics of MLLMs for autoregression.
For interpretability, Dex-AR \cite{bousselham2026dexar} computes layer-wise gradients of attention maps to trace token-level spatial explainability, distinguishing visually grounded tokens from purely linguistic priors.
AdaptInfer \cite{zhang2025adaptinfer} tracks evolving textual importance across layers to guide aggressive cross-modal pruning during the prefill stage.
DyRate \cite{liang2025dyrate} categorizes inputs into semantic chunks and notes that attention to visual tokens decreases as generation advances, and utilises this macroscopic analysis to prune redundant tokens.
While these works successfully monitor temporal attention shifts for spatial localization or token compression, the correspondence between attention routing and the semantic role of the token being generated remains underexplored.
Our work addresses this gap by relating attention dynamics to the output token semantics, causally isolating these temporal shifts to understand how models maintain fluency, ground themselves to the relevant modality, and execute task handoffs - without needing to isolate architectural sub-components.
Further, parallel to the monotonic decay of visual attention observed by \cite{liang2025dyrate}, we show that image relevancy is task-dependent and models attend to the visual context multiple times for multi-part tasks (\eg ChartQA).
Building on this, our test-time attention boosting strategy demonstrates that selectively amplifying modality-specific attention at critical decoding steps significantly improves overall multimodal performance.

\paragraph{Test-time attention intervention.}
A large body of research actively explores test-time attention interventions to improve LLMs and MLLMs.
\citet{wang2026ascd} show that contrastive decoding works 
by implicitly modulating internal attention, motivating explicit interventions.
While different approaches intervene at different computation steps, 
either on the logit~\cite{fazli2025mitigatingcaac, zhang2026adavboost, lyu2026pade}, or post-softmax~\cite{zhang2023pasta, kang2025seewhatyouaretold,gu2024llmsteer}, the non-trivial decision is \textit{where in the architecture} to intervene~\cite{zhang2023pasta, wang2026ascd, kang2025seewhatyouaretold, fazli2025mitigatingcaac, zhang2026adavboost}.
~\cite{liu2024paying_PAI, zhang2026adavboost} suppresses text while boosting the image.
A complementary axis targets individual tokens, either by selectively intervening on specific tokens~\cite{zhu2025ibd} or by manipulating causal masks to mitigate outliers~\cite{tang2025seeing}.
In contrast, we~\emph{globally} amplify the modality contextually relevant to the token at the right timestep.
The implicit reweighting
follows from the attention renormalization.
While~\cite{zhang2023pasta, kang2025seewhatyouaretold} report that global interventions degrade performance, they lack role-aware selectivity based on the tokens.
We address this gap by unifying global interventions with
modality-aware per-token amplification, and demonstrate the efficacy of this strategy.
\section{Conclusion}
\label{sec:conclusion}
Our work offered a novel temporal perspective on attention dynamics in multimodal LLMs.
We presented an approach to analyze attention patterns for an individual sample to the entire dataset.
Across multiple tasks and open-weight MLLMs, we observed consistent findings:
(i)~Models attended to relevant context, and disrupting attention at critical steps significantly degraded outputs, manifesting as language priors, cross-modal leakage, or denial.
(ii)~For multi-part responses, the instruction was revisited, and blocking attention at task handoff led to incomplete outputs.
(iii)~Previously generated tokens showed progressively increasing attention.
Upon disruption, this resulted in incomplete and repetitive outputs.
Finally, as an application, we leveraged the attention patterns to explain differences in model performance, and then improved performance through a simple attention boosting intervention.

\paragraph{Acknowledgements.}
MT and VG thank Adobe Research for supporting this project.
MT also thanks Google for Gemini credits and SERB SRG/2023/002544 for compute.

\clearpage
\bibliographystyle{plainnat}
\bibliography{bib/longstrings, bib/refs}

\appendix
\clearpage

\begin{center}
\Large \textbf{Appendix}
\end{center}

\titlecontents{section}
  [1.5em]
  {\small\bfseries\vspace{0.2ex}}
  {\contentslabel{1.8em}}
  {}
  {\titlerule*[0.5pc]{.}\contentspage}

\titlecontents{subsection}
  [3.3em]
  {\footnotesize}
  {\contentslabel{2.3em}}
  {}
  {\titlerule*[0.5pc]{.}\contentspage}

\startcontents[appendix]

\vspace{0.5cm}
{
\setlength{\parskip}{0pt}
\hypersetup{colorlinks=false, pdfborder={0 0 0}}
\printcontents[appendix]{}{1}[2]{}
}
\vspace{1cm}

In this appendix, we present
additional results in \cref{appendix:sec:results},
additional experimental setup and dataset/task details in \cref{appendix:sec:exp_setup},
some limitations, future work in \cref{appendix:sec:limitations}, and
all prompts used in this work in \cref{appendix:sec:prompts}.
Brief notes about the impact statement (\cref{appendix:subsec:impact_statement}) and compute resources (\cref{appendix:subsec:compute}) is also included.

\section{Additional Results}
\label{appendix:sec:results}

\subsection{Qualitative Results: Per-token Normalized Attention Plots}
\label{appendix:subsec:line_plots}

\begin{figure}[b!]
\small
\centering
A. Per-token attention to semantic chunks on a sample from the Fruit-Math task for \lovshort-0.5B.
\begin{minipage}{0.46\linewidth}
\includegraphics[width=\linewidth]{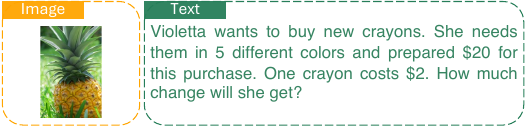}
\end{minipage}\hfill
\begin{minipage}{0.53\linewidth}
\includegraphics[width=\linewidth]{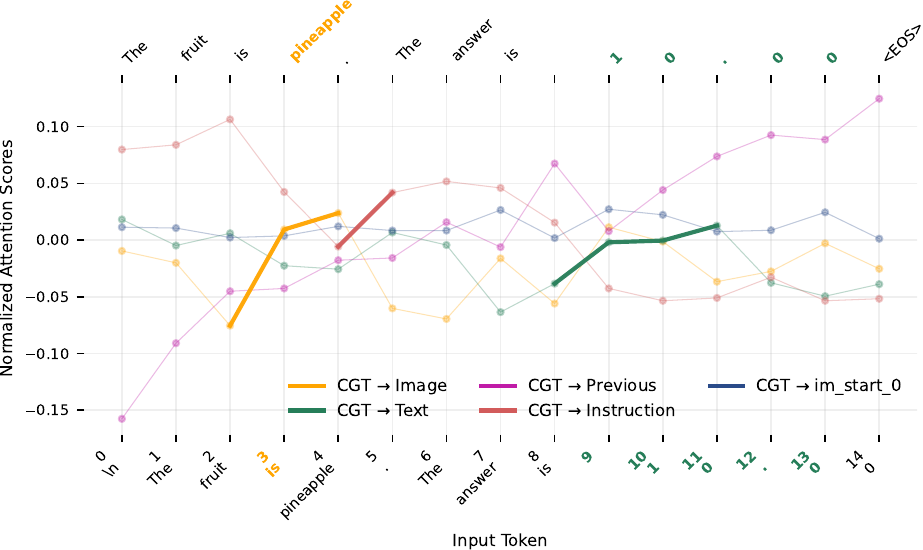}
\end{minipage}
B. Per-token attention to semantic chunks on a sample from the Fruit-Math task for \qvlshort-3B.
\begin{minipage}{0.46\linewidth}
\includegraphics[width=\linewidth]{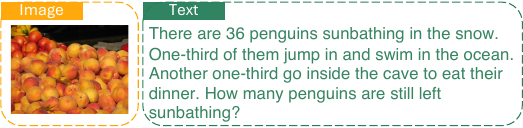}
\end{minipage}\hfill
\begin{minipage}{0.53\linewidth}
\includegraphics[width=\linewidth]{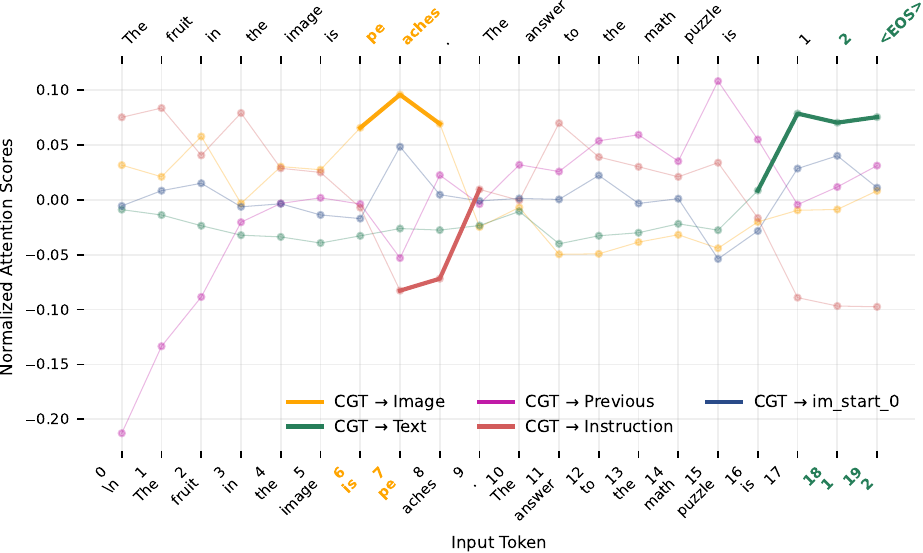}
\end{minipage}
C. Per-token attention to semantic chunks on a sample from the Fruit-Math task for \qvlshort-7B.
\begin{minipage}{0.46\linewidth}
\includegraphics[width=\linewidth]{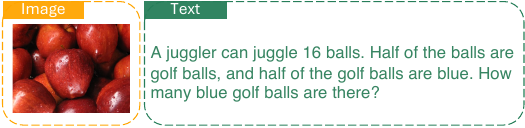}
\end{minipage}\hfill
\begin{minipage}{0.53\linewidth}
\includegraphics[width=\linewidth]{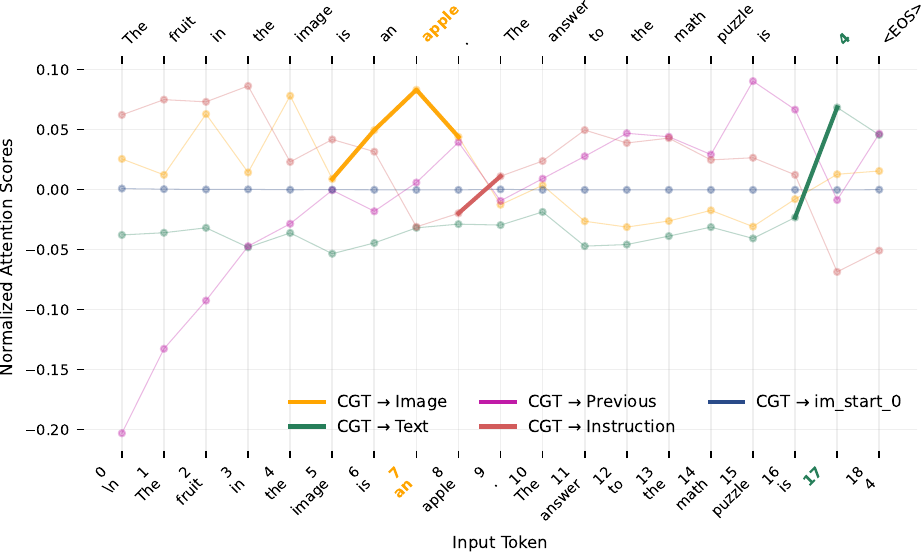}
\end{minipage}
\caption{Per-token normalized attention plots for all other models on the Fruit-Math task.
We observe similar attention patterns to the results presented in the main paper.}
\label{fig:additional_line_plots}
\end{figure}

Similar to \lov-7B's per-token attention plot in \cref{fig:teaser}E, we present additional plots for all other models (\lov-0.5B, \qvl-3B and \qvl-7B) in  \cref{fig:additional_line_plots}.
We pick different samples from the Fruit-Math task.
While local variances may exist within different models, global attention trends:
\imageOrange{image} peaking to generate the fruit concept (image task),
\textGreen{text} peaking to solve math (text task),
progressively increasing attention to \previousPink{previous} tokens, and
a flip-like increment on \instructionRed{instruction} during task-switch, continues to hold across all models.

Interestingly, results on individual samples also show that models frequently deviate from the prescribed output format.
We see \lovshort-0.5B go off script and say ``The fruit is'' instead of ``The fruit in the image is'', \etc.
This motivates the need and use of POS tagging to parse outputs that allow us to compile results across such generations (detailed in ~\cref{appendix:subsec:pos_tags}).

\subsection{Analyzing Attention Scores across Layer Groups}
\label{appendix: subsec: layer_grouped_analysis}
\begin{figure}
\centering
\includegraphics[width=\linewidth]{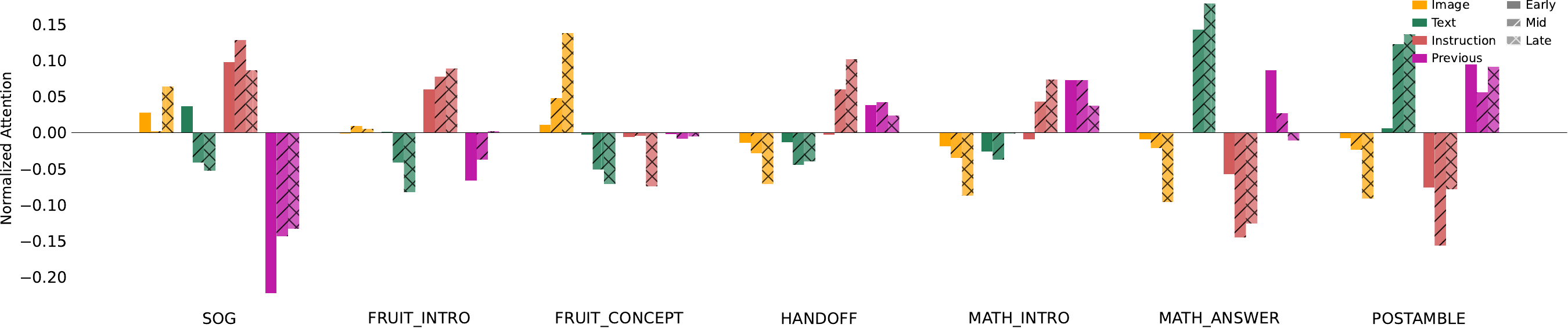}
\includegraphics[width=\linewidth]{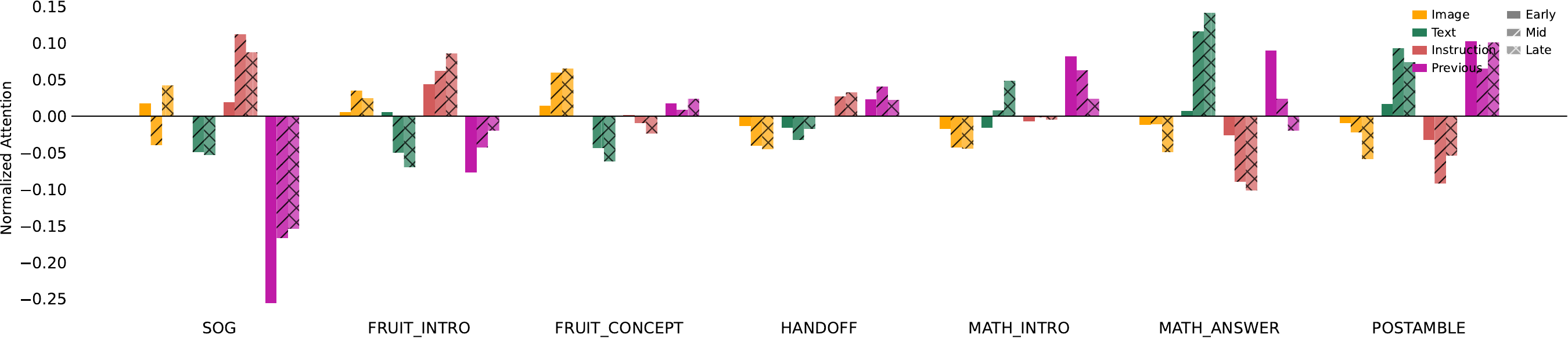}
\caption{Normalized attention analysis on the Fruit-Math task for \lovshort-7B (top) and \qvlshort-7B (bottom) models, grouped into early, middle, and late layers.}
\label{fig:layer_grouped_attention_plot}
\end{figure}

We conducted a stratified analysis of the Fr-Ma task, grouping attention into \textit{early}, \textit{middle}, and \textit{late} layers for both \lovshort-7B and \qvlshort-7B in \cref{fig:layer_grouped_attention_plot}.

This reveals a division of labor among the layers; the magnitude of attention varies significantly by depth, with late (and occasionally middle) layers driving the sharpest task-specific spikes (\eg the high attention to the \imageOrange{Image} at FRUIT\_CONCEPT, and to the \textGreen{Text} at MATH\_ANSWER).
Early layers exhibit a much flatter attention profile.
However, crucially, the directional trends remain consistent across depth.
During the FRUIT\_CONCEPT stage, early, mid, and late layers all exhibit positive relative attention to the \imageOrange{Image} chunk.
Because the layers operate in consensus on when to route specific modalities, differing primarily in the intensity of that routing, we find that global averaging does not obscure conflicting layer-wise mechanisms.
Instead, it serves as a robust, macroscopic summary of the model’s overall generative orchestration - without introducing model-specific nuances, and predominantly reflecting the strong routing signals of the later, predictive layers.

\subsection{Attention Trends on Fr-Ma for Additional Models}
\label{appendix:subsec:attention_trends}

\begin{figure}[h!]
\centering
\includegraphics[width=0.49\linewidth]{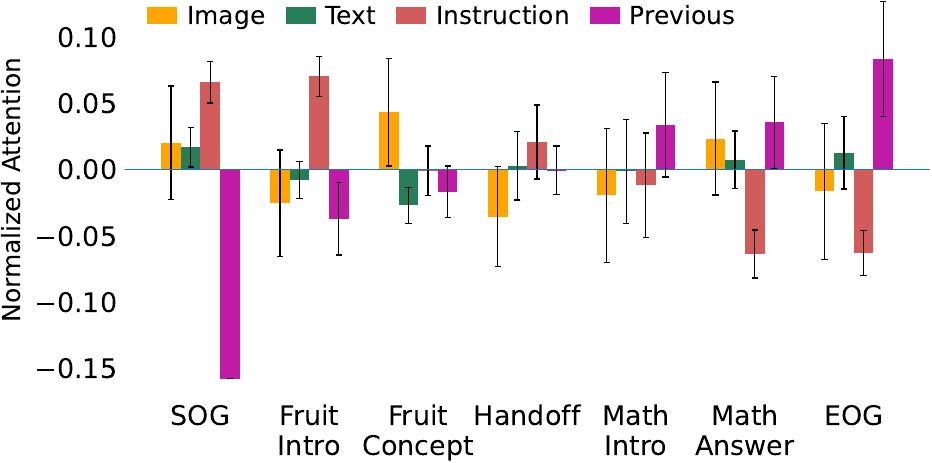}
\includegraphics[width=0.49\linewidth]{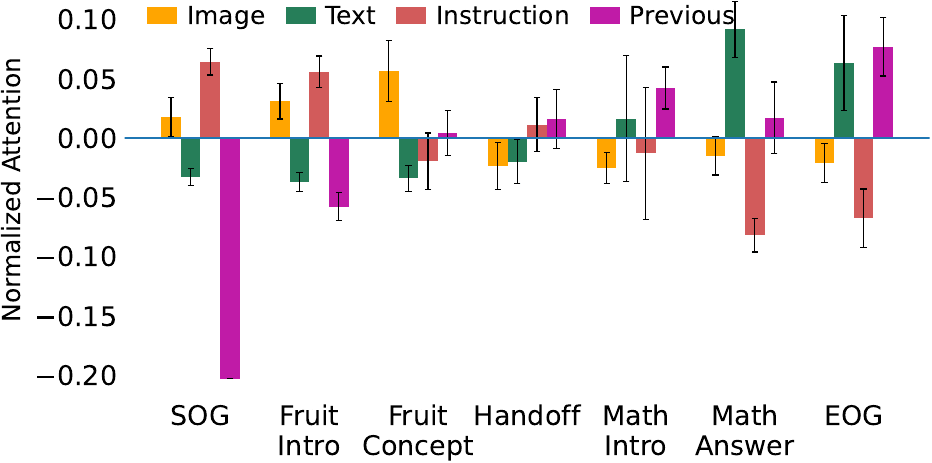}
\caption{Attention trends for \lov-0.5B (left) and \qvl-7B (right) across output tags on the Fruit-Math task.}
\label{fig:attention_trends_qvl7B_lov0.5B}
\end{figure}

Similar to the results in the main paper \cref{fig:attention_mathfruit}, we present attention trends for the remaining two models, \lov-0.5B and \qvl-7B in \cref{fig:attention_trends_qvl7B_lov0.5B}.
\qvl-7B shows the same trends as models reported in the paper.

Being a small model, \lov-0.5B tends to show higher variance.
However, average attention still peaks at the \imageOrange{image} when decoding \textit{Fruit Concept}, at the \instructionRed{instruction} during \textit{Handoff}, and at the math puzzle when solving it (\textit{Math Answer}).
Interestingly, the \textGreen{text} peak is low, perhaps indicative of the model's struggles in solving basic math questions with $\sim$4\% accuracy.

\subsection{Attention Patterns in Text-only LLMs}
\label{appendix:subsec:llm_only_experiment}
LLMs are adapted for visual inputs by incorporating visual embeddings and fine-tuning them.
A natural question is: Do LLMs show similar attention patterns
when performing two text tasks?
We present our approach of evaluating LLMs in this section.

\paragraph{Math-Sport task setup.}
To conduct similar multi-task analysis on LLM-only models, we formulate the \textbf{Math-Sport} task.
This is a \emph{text-only} diagnostic dataset.
Here, we pair math puzzles with sport excerpts (\cref{fig:sport_data}) that require identifying the sport given a paragraph related to it.
This enables us to draw parallels between text-only LLMs and vision MLLMs as in the main paper.
We prompt LLMs to solve the math puzzle and identify the sport.
The data samples with categories are shown in \cref{fig:sport_data},
the sport-text excerpt creation prompt is present in \cref{fig:gemini_sport_text_creation_prompt}, 
and the task prompt is specified in \cref{fig:generation_prompt_MaSp}.

\paragraph{Experiment and observations.}
We evaluate Qwen2-7B \cite{yang2024qwen2llm}, the base LLM used by LOV-7B \cite{li2025llavaonevision},
on the Math-Sport task, and find that
attention patterns in LLMs mimic those of MLLMs when solving two disjoint text tasks.

\begin{figure}[h]
\centering
\includegraphics[width=0.55\linewidth]{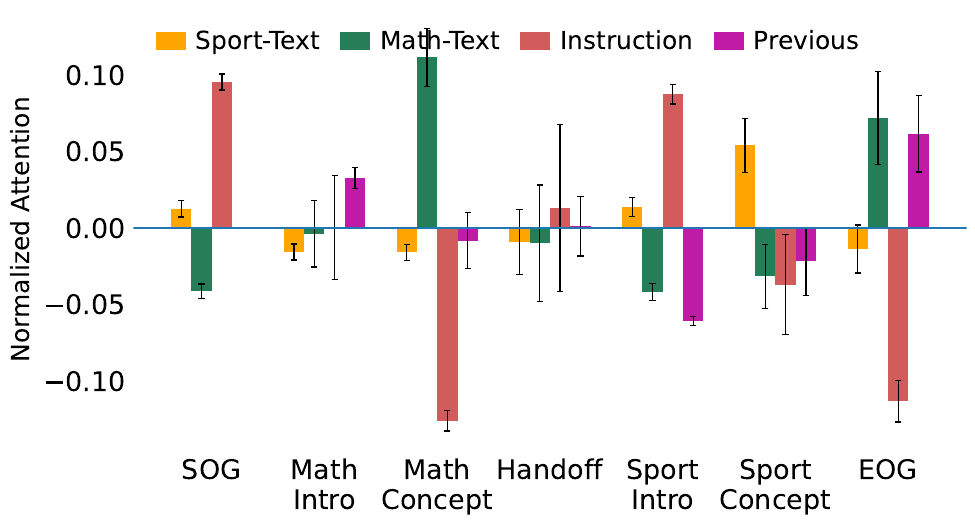}
\caption{Attention trends for Qwen2-7B LLM (the base LLM of LOV-7B) on the Math-Sport task.}
\label{fig:q2llm_math_sport_bar_plot}
\end{figure}

\cref{fig:q2llm_math_sport_bar_plot} shows similar tendencies: attention spikes to sport text when predicting the \textit{Sport Concept} and math puzzle text when generating \textit{Math Answer}.
Other attributes, such as increasing attention to previously generated tokens and higher attention to the instruction during Handoff are also observed.

This hints that our findings could generalize to base LLMs, and it is likely that MLLMs derive these features from the base LLM during multimodal training.

\subsection{Qualitative Results on Fruit Prediction: Blocking Information Flow from Image}
\label{appendix:subsec:blocking_qualitative_fruit}

We present some qualitative results for the impact of blocking attention to image chunks when generating the \textit{Fruit Concept}.
As presented in \cref{subsec:results:attention_blocking}, while accuracy drops for all models, the reasons for this drop are interesting.

\begin{figure}[h!]
\centering
\includegraphics[width=0.8\linewidth]{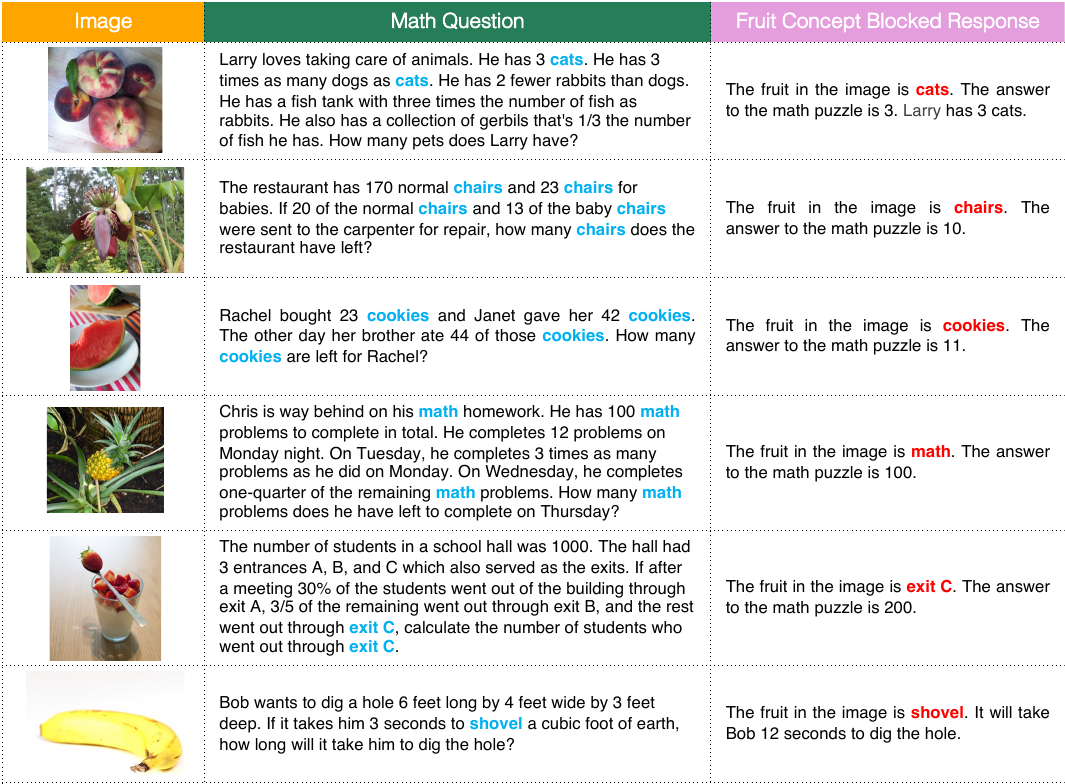}
\caption{\lov-0.5B exhibits strong cross-modal information leakage where the fruit predictions are simply nouns borrowed from the math question. We visualize some examples.}
\label{fig:ov05b_leakage}
\end{figure}

\paragraph{Examples of cross-modal information leakage in \lovshort-0.5B.}
\cref{fig:ov05b_leakage} shows multiple examples of how nouns mentioned in the math puzzle leak to the name of the identified fruit.
This results in weird generations such as the fruit in the image is ``cats'' / ``cookies'' / ``math'', and even very specific noun-phrases such as ``exit C''!

\begin{figure}[h!]
\centering
\includegraphics[width=0.8\linewidth]{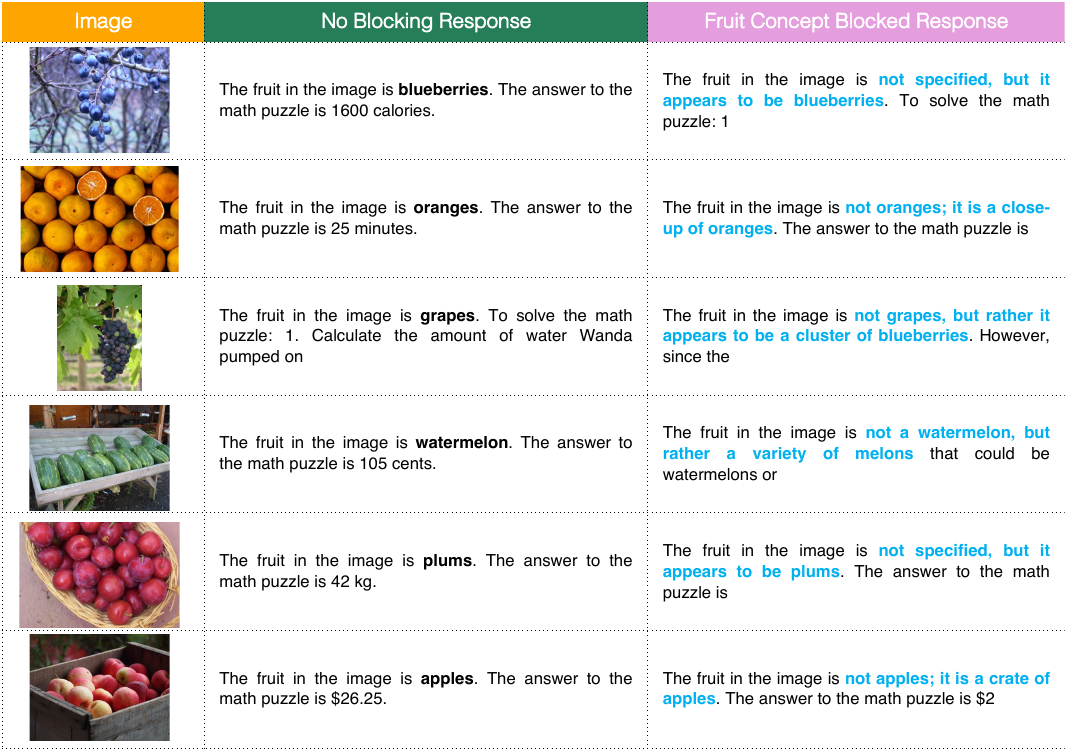}
\caption{\qvl-3B shows mixed results with examples of denial, reliance on language prior, and some examples of recovery where the model ends up predicting the fruit correctly.
We visualize several recovery examples.}
\label{fig:rescue_qvl3b}
\end{figure}

\paragraph{Examples of recovery in \qvlshort-3B.}
\cref{fig:rescue_qvl3b} presents some examples of recovery.
We observe grammatically confusing patterns such as repetition of the same fruit ``not apples; it is a crate of apples'',
use of additional placeholder words to produce the correct answer ``not specified, but appears to be plums'', or 
an unnecessary attribute/detail that the model identifies ``not oranges, it is a close-up of orranges''.

\begin{figure}[h!]
\centering
\includegraphics[width=0.8\linewidth]{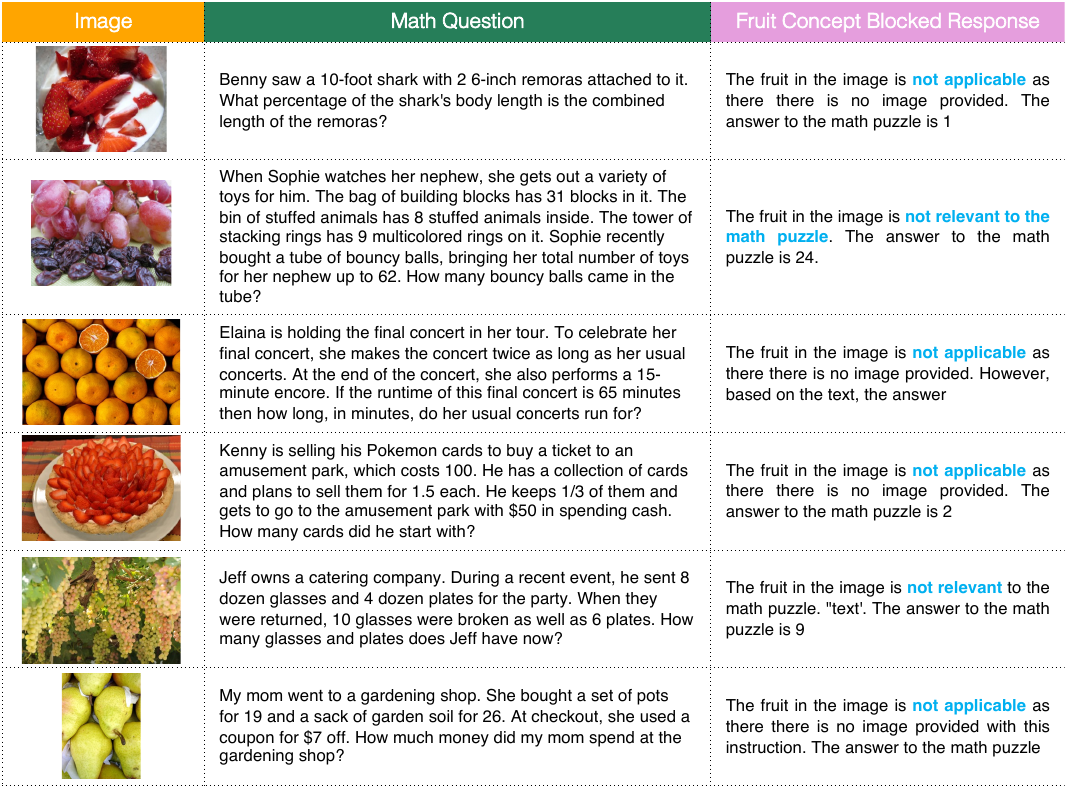}
\caption{\qvl-7B is able to understand that the image does not exist (in this case, due to blocking), and produces outputs to this effect.
We show some examples from the roughly 44\% of samples that show such behavior.}
\label{fig:denial_qvl7B}
\end{figure}

\paragraph{Examples of denial exhibited by \qvlshort-7B.}
\cref{fig:denial_qvl7B} presents cases where the model claims that the fruit in the image is ``not applicable'' or ``not relevant''.
Note, even though blocking is applied only at one step, the model continues to believe that the image does not exist and we see this in the 3rd example:
``The fruit in the image is not applicable as there is no image provided.''.
Interestingly, here, the model also fails to switch tasks and answer the math puzzle.

\subsection{Qualitative Results: Other Types of Blocking}
\label{appendix:subsec:blocking_other}

Similar to results of blocking image information, we show qualitative results for blocking other semantic chunks.
While all models show similar behavior, we present examples across models.

\begin{figure}[h!]
\centering
\includegraphics[width=0.8\linewidth]{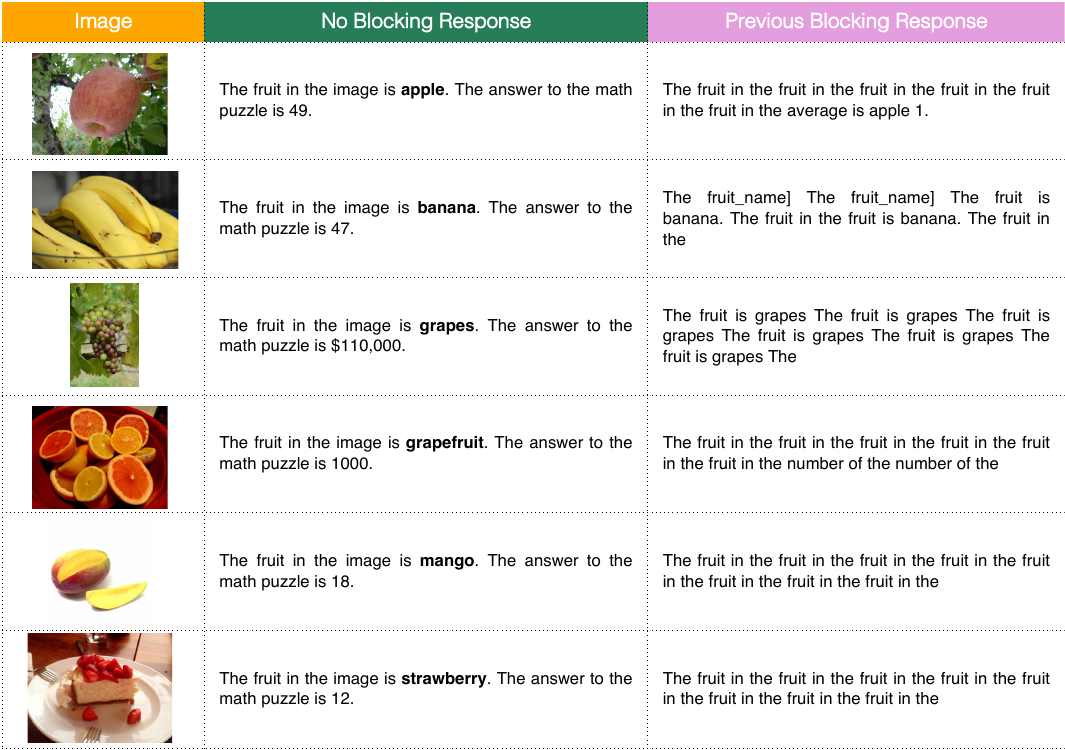}
\caption{Outputs produced by \lov-7B when attention to previously generated tokens is blocked.
We see that the output sentence collapses and there is repetition or premature end of generation.}
\label{fig:prev_block}
\end{figure}

\paragraph{Blocking previously generated tokens.}
\cref{fig:prev_block} shows that attention to previously generated tokens is crucial for the model to generate appropriate responses.
Among the examples, we see that most outputs stutter and break after generating a few words such as ``The fruit in the ...'' and generate repetitive outputs.
In the third example, we see \lovshort-7B generate the correct answer (grapes).
However, it is unable to stop repeating ``the fruit is grapes the fruit is grapes [...]'' and switch to the next task.
Nevertheless, as quantified in \cref{tab:attention_blocking_other}, \qvl{} shows higher resilience towards blocking information flow from previously generated tokens.

\begin{figure}[h!]
\centering
\includegraphics[width=0.8\linewidth]{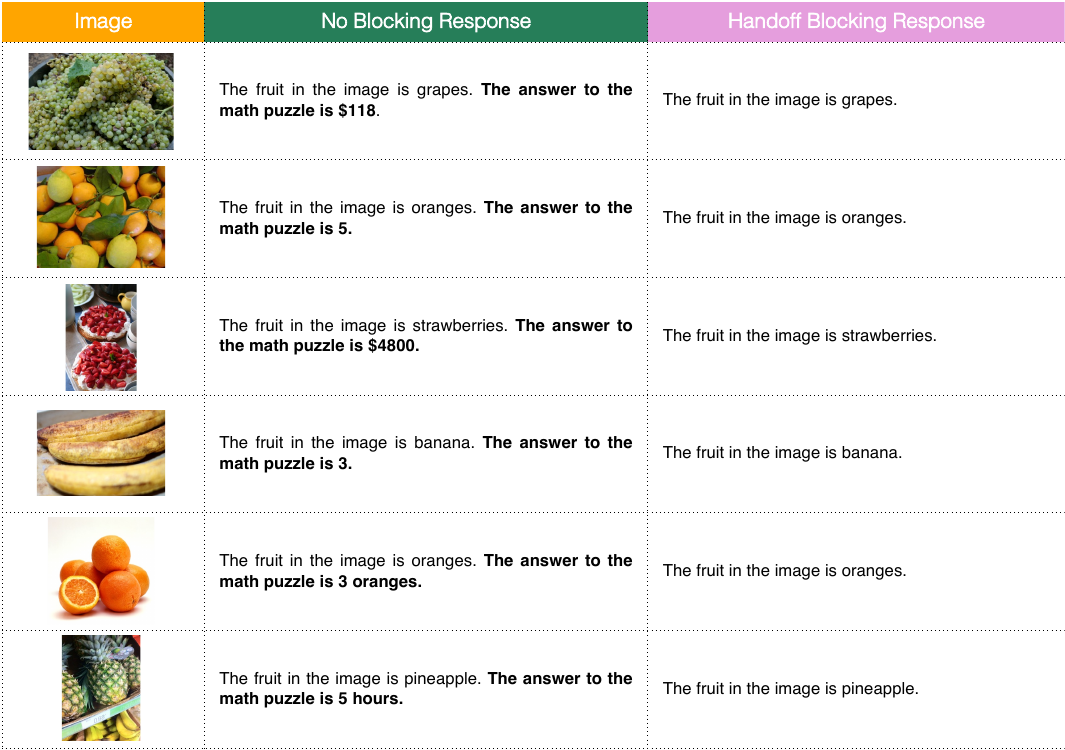}
\caption{Outputs produced by \qvl-3B when attention to the instruction token is blocked during \textit{Handoff}, or switching from the Fruit to the Math task.
Compared to the case without blocking, models generate responses for the fruit identification task, but are unable to generate the math answer and end outputs prematurely.}
\label{fig:handoff_block}
\end{figure}

\paragraph{Blocking instruction when generating the handoff token.}
Attention to the instruction chunk peaks when switching between tasks.
\cref{fig:handoff_block} shows some qualitative outputs of blocking attention to the instruction chunk at the \textit{Handoff} token.
We see that the models fail to generate any outputs for the Math task (or even acknowledge that it exists).

\subsection{Boost Factor Analysis and $\gamma$ Computation}
\label{appendix:subsec:boost_factor_gamma}
\begin{figure}
\centering
\includegraphics[width=0.6\linewidth]{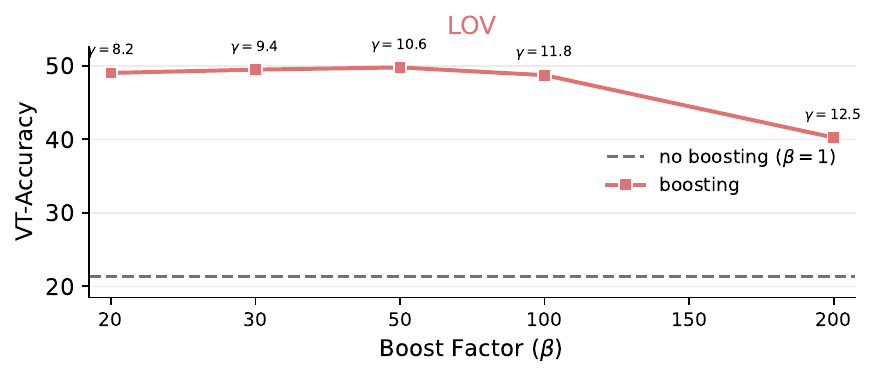}
\includegraphics[width=0.6\linewidth]{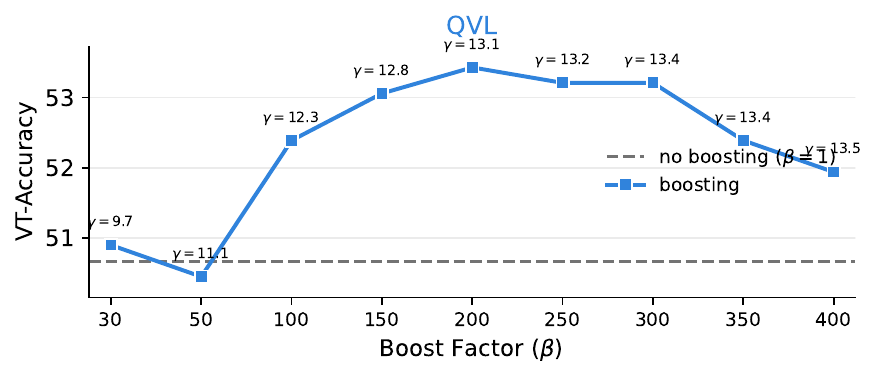}
\caption{Boosting analysis across wide range of $\beta$ tested for \lovshort-7B and \qvlshort-7B.}
\label{fig:boosting_sensitivity_plot}
\end{figure}

We present an analysis on the attention boost factor discussed in \cref{subsec:results:attention_boosting}.
As can be seen in \cref{fig:boosting_sensitivity_plot}, VT-Acc remains largely stable across a broad range of boost factors.
For LOV, performance peaks at 49.8 with $\beta = 50$ and stays within 1.1 points of this maximum for $\beta \in [20,100]$.
Similarly, QVL varies by less than 1 point between $\beta = 30$ and $\beta = 400$, with a shallow optimum around 150-300.
These results indicate that the benefits of attention boosting are not tied to a narrowly tuned hyperparameter choice; rather, the method exhibits a wide plateau of strong performance, suggesting low sensitivity to the exact value of $\beta$ within a broad operating region.

The $\gamma$ values in \cref{tab:attn_boost} for the VSR task are computed as follows,
For \lovshort-7B, the $p_i$ calculated from the recorded attention values on the VSR task is $0.0756$.
With $\beta = 50$, this results in $Z_i = (1 - 0.0756) + 50 \times (0.0756) = 4.7044$, and a $\gamma_i = 50 / 4.7044 = 10.62$.
Similarly, for \qvlshort-7B, with $p_i = 0.0718$ and $\beta = 200$, results in $\gamma_i = 13.08$.

\subsection{Qualitative Results: Attention Boosting Intervention}
\label{appendix:subsec:qual_boosting_intervention}

\begin{figure}
\centering
\includegraphics[width=0.8\linewidth]{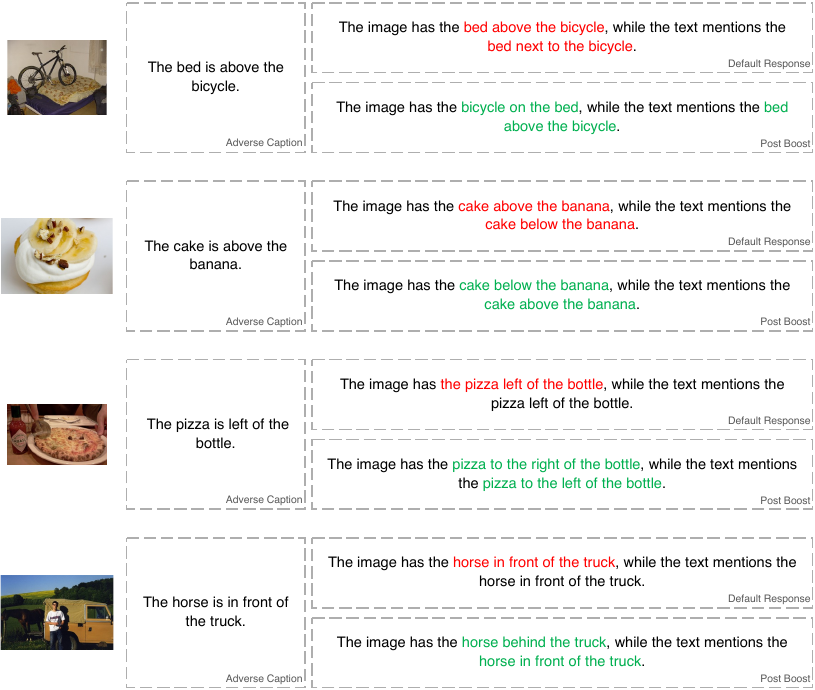}

\caption{Attention intervention qualitative samples on \lovshort-7B on the VSR task.}
\label{fig:lov_7b_attn_boosting}
\end{figure}

\begin{figure}
\centering
\includegraphics[width=0.8\linewidth]{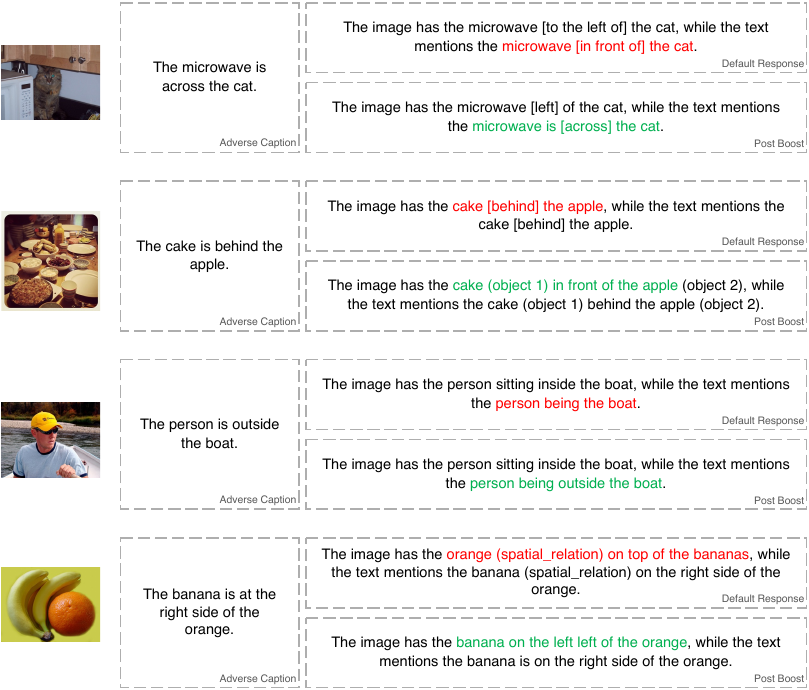}

\caption{Attention intervention qualitative samples on \qvlshort-7B on the VSR task.}
\label{fig:qvl_7b_attn_boosting}
\end{figure}

We present qualitative results of boosting intervention on \qvl-7B and \lov-7B models.  The intervention improves both image-based and text-based spatial reasoning accuracy in \lov, consistent with \cref{tab:attn_boost}. \qvl has mixed gains, with certain answers now correctly answering the text spatial answer, and sometimes both, image and text, improve. Samples reported in \cref{fig:lov_7b_attn_boosting} and \cref{fig:qvl_7b_attn_boosting}.

\subsection{Significance Testing of Attention Trends}
\label{appendix:subsec:significance_testing}
\begin{table}[b]
\small
\centering
\tabcolsep=3pt
\caption{
Attention specialization across modalities.
Positive $\Delta$ indicates stronger attention for the first group
in each comparison. All reported differences are statistically
significant under Welch's t-test ($p<0.05$).
}
\label{tab:significance_testing}

\begin{tabular}{llccccc}
\toprule
Task & Comparison & Modality
& \lovshort-0.5B
& \lovshort-7B
& \qvlshort-3B
& \qvlshort-7B \\
\midrule

\multirow{3}{*}{Fr-Ma}

& FC $>$ FI
& Image
& +0.063
& +0.061
& +0.048
& +0.030 \\

& MA $>$ MI
& Text
& +0.009
& +0.130
& +0.095
& +0.075 \\

& HO $>$ FC
& Instruction
& --
& +0.044
& --
& +0.012 \\

\midrule

\multirow{2}{*}{ChartQA}

& Q1\_A $>$ Q1\_I
& Image
& --
& +0.106
& --
& +0.085 \\

& Q1\_A $>$ Q1\_I
& Text
& --
& +0.019
& --
& +0.003 \\

\bottomrule
\end{tabular}
\vspace{-3mm}
\end{table}

We quantify modality-specific attention specialization using two-sided Welch's t-tests between semantically contrasted groups.
For Fr-Ma, we compare Fruit Concept (FC) against Fruit Intro (FI)
for image attention, Math Answer (MA) against Math Intro (MI)
for text attention, and Handoff (HO) against FC
for instruction attention in \cref{tab:significance_testing}.

Across models, FC consistently allocates higher image attention
than FI, while MA allocates higher text attention than MI.
For ChartQA, answer-bearing tokens (Q1\_A) exhibit substantially
stronger image attention than introductory tokens (Q1\_I),
with comparatively weak separation in text attention.

Reported values indicate mean attention differences
($\Delta$ mean), where positive values denote stronger attention
for the first group in each comparison.
All reported differences are highly statistically significant ($p<0.05$).


\section{Experiment Setup Details}
\label{appendix:sec:exp_setup}

\subsection{Sequence of Input Context Chunks}
\label{appendix:subsec:chunks}

\begin{figure}[h!]
\centering
\includegraphics[width=\linewidth]{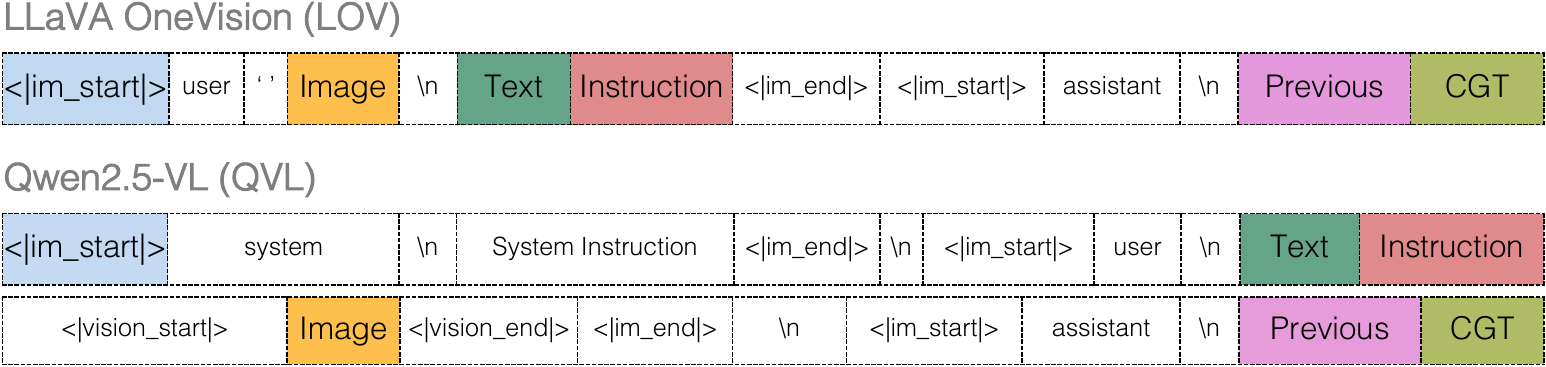}
\caption{Sequence of input context chunks for \lov{} (top) and \qvl{} (bottom).
Note their key differences in how image tokens may appear before/after the text and instruction tokens.
}
\label{fig:input_chunks}
\end{figure}

Our work focuses on studying attention patterns from the currently generating token (CGT) to various context chunks.
To do so, it is important to know the exact sequence of tokens.
\cref{fig:input_chunks} shows the order of tokens for two popular MLLM families that we evaluate in our work: \lov{} and \qvl{}.

Some differences are apparent.
First, we see that token order is not simply image, concatenated with text and/or instruction, followed by the model's generation.
Even \lov{} uses tags and indicators such as \texttt{im\_start} or \texttt{im\_end} to mark conversational boundaries between the \textit{user} and the \textit{assistant} (model).

We also observe that \lov{} adopts an image before text and instruction sequence order, while \qvl{} prefers to send in text and instruction tokens before the image.
Due to the use of causal or masked attention in the decoder, the order of the tokens determine how information flows in an MLLM and can possibly change how the models are interpreted.


\subsection{Dataset Details}
\label{appendix:subsec:dataset}

\begin{figure}[h]
\centering
\includegraphics[width=0.4\linewidth]{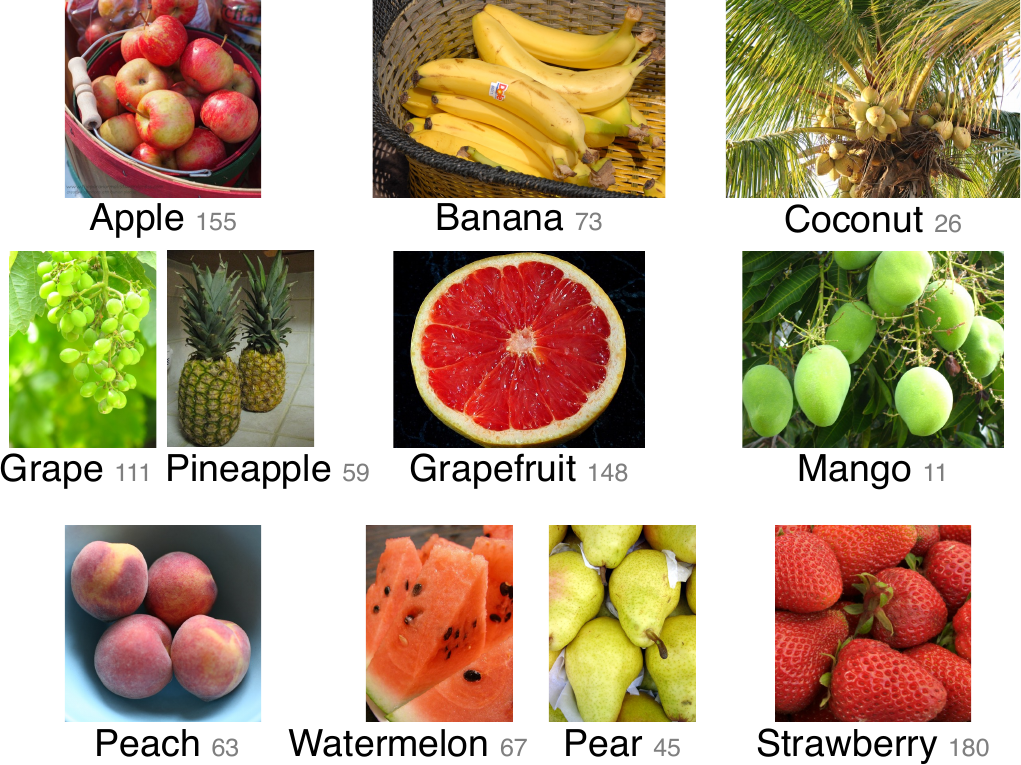}
\caption{Fruit categories in our fruit dataset.}
\label{fig:fruit_data}
\end{figure}

Our dataset consists of 11 fruit categories, which are shown in \cref{fig:fruit_data}, along with their respective counts.

\begin{figure}[h!]
\centering
\includegraphics[width=0.8\linewidth]{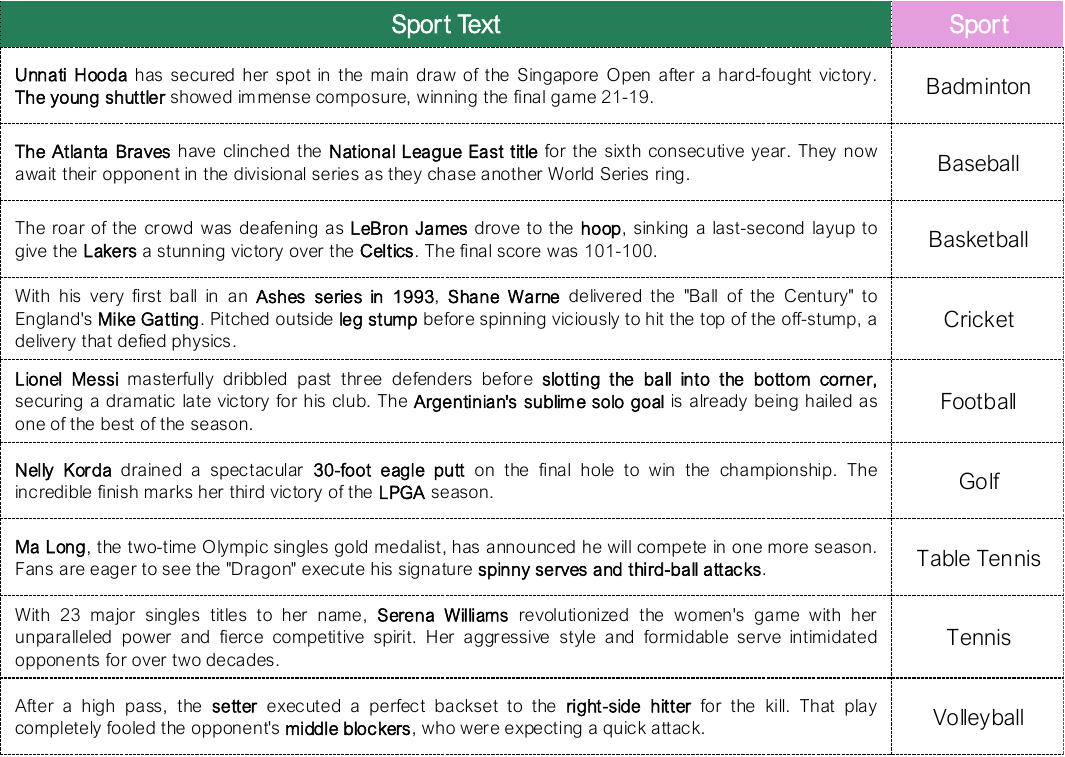}
\caption{Examples of sport themed excerpts generated by Gemini for each of the 9 sport classes used in our experiments for Math-Sport (LLM text-only results in \cref{appendix:subsec:llm_only_experiment}).
While the name of the sport is not mentioned in the excerpt, they are fairly easy to guess based on the descriptions.}
\label{fig:sport_data}
\end{figure}

\paragraph{Sport categories.}
\cref{fig:sport_data} presents some example text excerpts used in our sport identification task for LLM text-only results in \cref{appendix:subsec:llm_only_experiment}.
These are generated by Gemini 2.5 Pro (see prompt in \cref{fig:gemini_sport_text_creation_prompt}).


\subsection{POS Tags}
\label{appendix:subsec:pos_tags}
\begin{figure}
\centering

\begin{minipage}{0.48\linewidth}
\centering
\includegraphics[width=\linewidth]{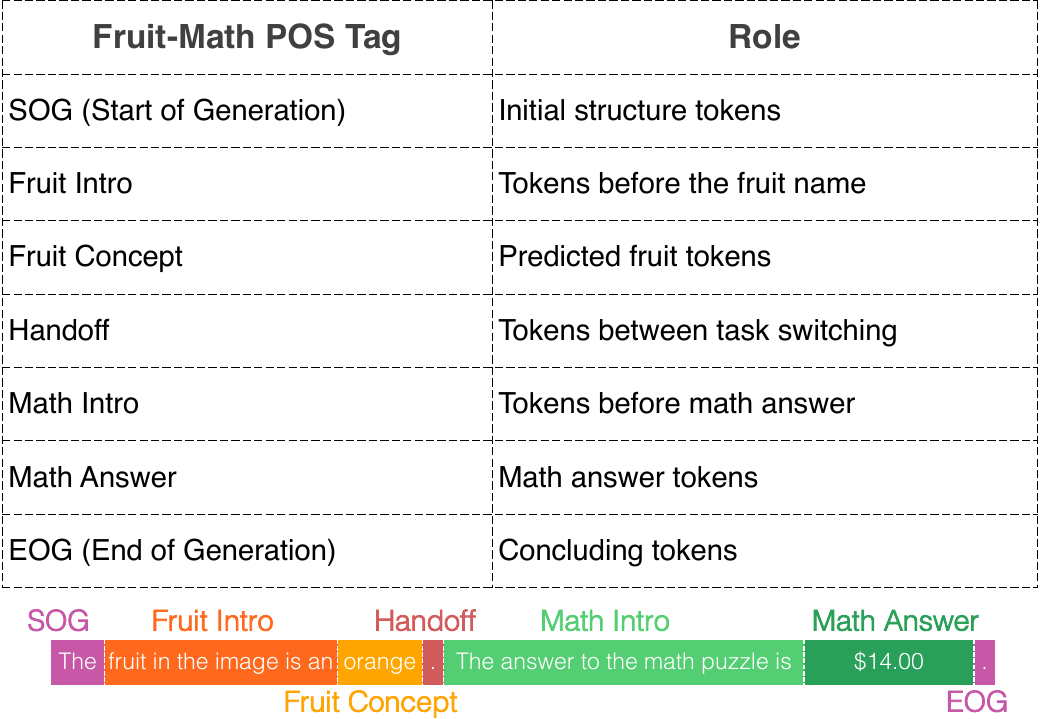}
\end{minipage}
\hfill
\begin{minipage}{0.48\linewidth}
\centering
\includegraphics[width=\linewidth]{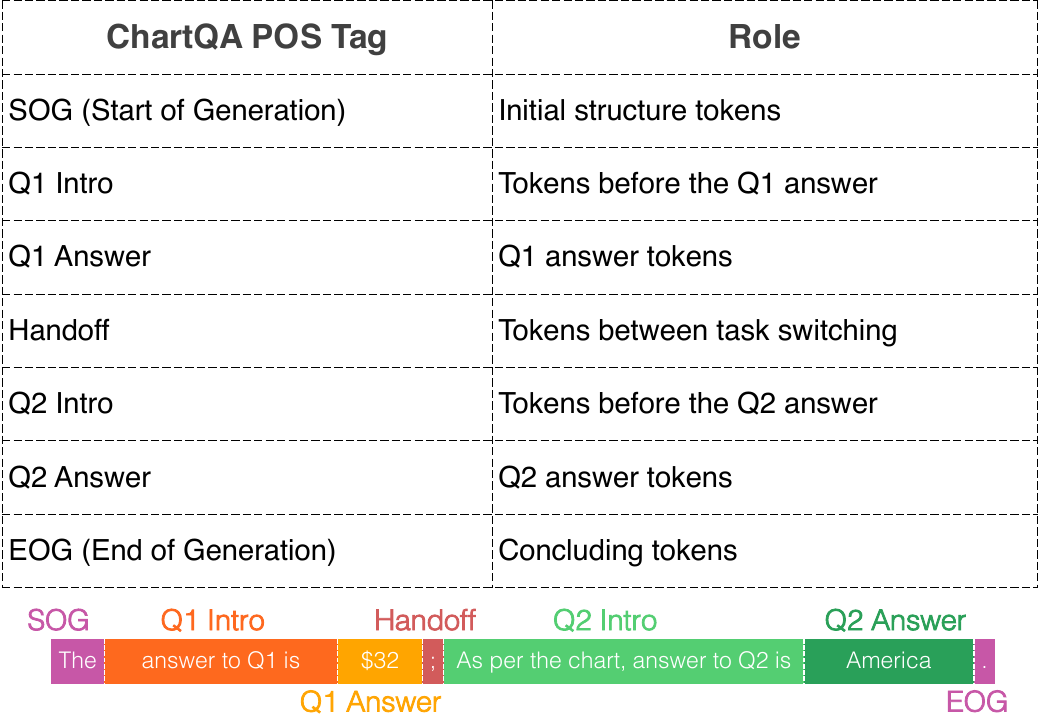}
\end{minipage}

\vspace{0.5cm}

\begin{minipage}{0.48\linewidth}
\centering
\includegraphics[width=\linewidth]{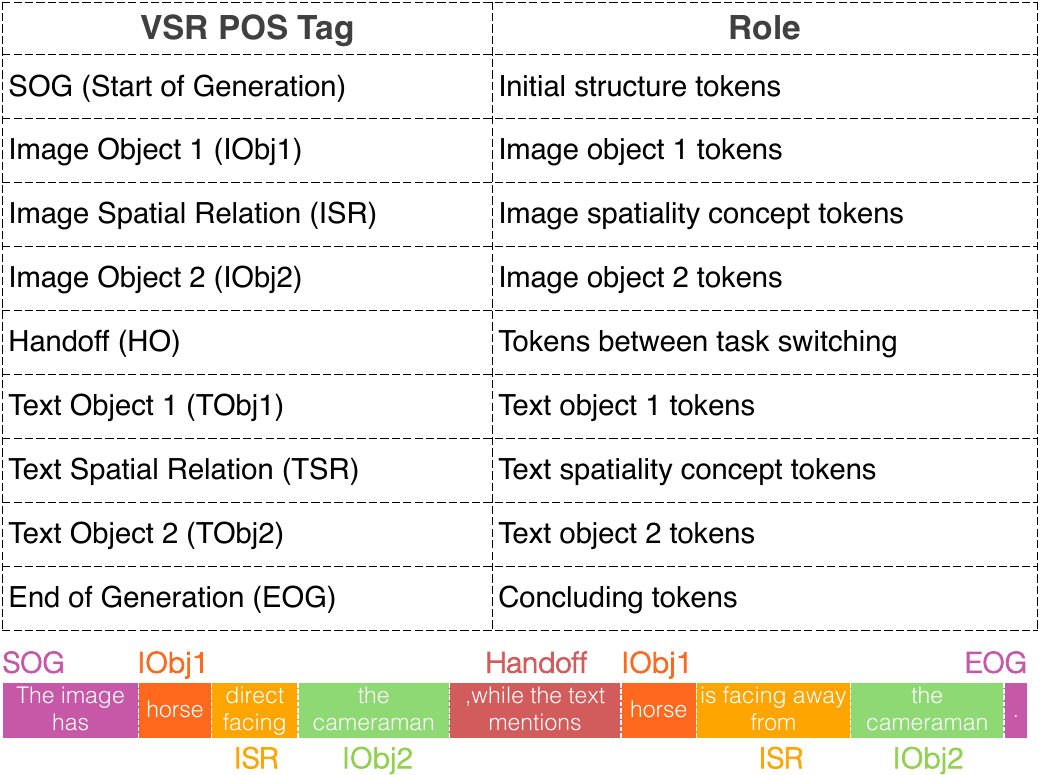}
\end{minipage}

\caption{POS tagging for Fruit-Math, ChartQA, and VSR tasks. Dataset-level findings reported in \cref{sec:results}.}
\label{fig:pos_tags_figure}
\end{figure}

To compile dataset-level trends (reported as bar plots in the paper), we rely on automated POS tagging to group token roles and aggregate across these roles.
Different tasks follow slight tagging variations, and we present them in \cref{fig:pos_tags_figure}.

\subsection{Details about When to Block}
\label{appendix:subsec:when_block}

We present details on the aspect of \textbf{when} information flow is blocked from various semantic chunks to CGT.
First, let us consider the specified response format:
``The fruit in the image is [fruit\_name]. 
The answer to the math puzzle is [numeric\_answer].''
If the fruit name is \textit{apple} and the math answer is \textit{7}, we can write the expected sequence of output tokens as:
\begin{eqnarray*}
(
y_{1} &=& \text{``The''}, \,\,\,
y_{2} = \text{``fruit''}, \,\,\,
y_{3} = \text{``in''}, \,\,\,
y_{4} = \text{``the''}, \,\,\,
y_{5} = \text{``image''}, \,\,\,
y_{6} = \text{``is''}, \,\,\, \\
y_{7} &=& \text{``apple''}, \,\,\,
y_{8} = \text{``.''}, \,\,\,
y_{9} = \text{``The''}, \,\,\,
y_{10} = \text{``answer''}, \,\,\,
y_{11} = \text{``to''}, \,\,\,
y_{12} = \text{``the''}, \,\,\, \\
y_{13} &=& \text{``math''}, \,\,\,
y_{14} = \text{``puzzle''}, \,\,\,
y_{15} = \text{``is''}, \,\,\,
y_{16} = \text{``7''}, \,\,\,
y_{17} = \lbrace\text{EOS}\rbrace
) \, .
\end{eqnarray*}
While we could assume deterministic outputs and block at specific indices (\eg~$y_7$ for fruit name), as shown earlier, models deviate from the prescribed format.
Note, these challenges are not applicable to previous works that study attention blocking only for the first token $y_1$.

Now, at step $t$, we know that autoregressive generation takes $y_{t-1}$ as input to produce $y_t$.
Thus, we can consider the input that is fed to the model to calculate attention and generate $y_t$ to determine whether information should be blocked.
For example, for the \textit{Fruit Concept} blocking experiments, we wait until the previously generated token $y_{t-1} = \text{``is''}$ as we know that the model is instructed to generate the fruit name next at $y_t$.
Thus, we block information flow here.
Similarly, for \textit{Handoff} and instruction chunk blocking experiments, we identify the moment when $y_{t-1} = \text{``.''}$ (period) as it indicates that the next word will start the next sentence corresponding to the other task (math task in Fruit-Math).
Both are examples of blocking at a single step.

When we perform experiments for blocking at \textit{Fruit Intro}, this consists of multiple steps.
As the model may not adhere to the instruction we block attention to CGT until the input token is $y_{t-1} = \text{``is''}$ as the next token ($y_t$) is now likely to be the \textit{Fruit Concept}.
Note, we can also pose step index constraints to this process such as $t \in [4, 8]$ to ensure that the specific chosen word, ``is'' in this case, corresponds to the first appearance in our instruction.

Our rule-based blocking approach as defined above is scalable and allows us to obtain a holistic dataset level perspective on how attention patterns evolve across time.

\section{Limitations and Future Work}
\label{appendix:sec:limitations}
We identify a few limitations of our work that could guide future explorations.

\begin{itemize}[topsep=0pt, leftmargin=*]
\item Generation-centric evaluations are difficult to scale, both in terms of systematic evaluation and causal intervention.
While specifying a response format in the instruction and using Gemini to assist in tagging and grouping helps, complete free-form in-the-wild generation remains to be explored.

\item Attributing \emph{why} specific models exhibit distinct behaviors remains challenging.
For instance, \qvl{} frequently denies the presence of visual content under attention disruption, whereas \lov{} models do not.
Isolating the architectural aspect or training process responsible for such differences is not trivial.

\item Our work demonstrated the effectiveness of role-aware attention boosting in a global setting across all layers and heads. However, several factors remain underexplored, including decoding strategies and the optimal boost factor for different models. A systematic investigation of these aspects would require an independent study and is therefore beyond the scope of the current work.

\item Attention analysis requires inference in \textit{eager} attention mode, which materializes the full attention matrix. While it enables access to global attention statistics, it's significantly slower and compute-intensive than optimized implementations like FlashAttention~\cite{dao2022flashattention}, which are non-trivial to analyse.
\item Finally, our notion of multimodal interpretability is limited to vision-language interactions, leaving other modalities unexplored, and forms an interesting future pursuit.
\end{itemize}

\section{Prompts}
\label{appendix:sec:prompts}
\subsection{Sport Text Creation Prompt}
\label{appendix:subsec:sport_text_prompt}

\begin{figure}[h!]
\centering
\noindent\begin{minipage}{\textwidth}
\mdfsetup{%
middlelinewidth=1pt,
backgroundcolor=yellow!7,
innerleftmargin=0.5cm,
innerrightmargin=0.5cm,
font=\small,
roundcorner=15pt}
\begin{mdframed}
\vspace{0.2em}

Your task is to generate 12 text snippets for a text classification dataset.

\medskip

Rules:
\begin{itemize}[leftmargin=*, nosep]
    \item The text must be about \{sport\}.
    \item The text MUST NOT contain the word ``\{sport\}''.
    \item The text MUST imply the sport by using sport specific jargon - including actions, matches, name of the key players and athletes, etc.
    \item Each snippet should be 2-3 lines long, and snippets should be diverse in text content.
    \item The snippets should read like news clips.
\end{itemize}

\medskip

You may also use recent or older news on the internet to make this.

\vspace{0.2em}
\end{mdframed}
\end{minipage}
\caption{Gemini sport text creation prompt for various sports.}
\label{fig:gemini_sport_text_creation_prompt}
\vspace{-1em}
\end{figure}

The prompt used to create the sport themed experts for the base LLM (Sport-Math) task is presented in \cref{fig:gemini_sport_text_creation_prompt}.
This allows us to generate samples without contamination of the sport name (also verified manually).

\subsection{MLLM Task Prompts}
\label{appendix:subsec:mllm_task_prompts}

\begin{figure}[h!]
\centering
\noindent\begin{minipage}{\textwidth}
\mdfsetup{%
middlelinewidth=1pt,
backgroundcolor=yellow!7,
innerleftmargin=0.5cm,
innerrightmargin=0.5cm,
font=\small,
roundcorner=15pt,
frametitlebackgroundcolor=cyan!30,
frametitlefont=\bfseries\small,
frametitlerule=true,
frametitlerulewidth=0.8pt,
frametitleaboveskip=0.3em,
frametitlebelowskip=0.5em}
\begin{mdframed}
\vspace{0.2em}
\strut \textbf{\small Fruit-Math Prompt}
\vspace{0.4em}
\hrule
\vspace{0.6em}

`Text': [[text\_fill]]

\medskip

First, identify the fruit in the image provided and then answer the math puzzle in `Text'.

\medskip

Generate response in the format: The fruit in the image is [fruit\_name]. The answer to the math puzzle is [numeric\_answer].

\vspace{0.2em}
\end{mdframed}
\end{minipage}
\caption{MLLM generation prompt for Fruit-Math task.}
\label{fig:generation_prompt_FrMa}
\vspace{-1em}
\end{figure}
\begin{figure}[h!]
\centering
\noindent\begin{minipage}{\textwidth}
\mdfsetup{%
middlelinewidth=1pt,
backgroundcolor=cyan!10,
innerleftmargin=0.5cm,
innerrightmargin=0.5cm,
font=\small,
roundcorner=15pt,
backgroundcolor=yellow!7,
frametitlefont=\bfseries\small,
frametitlerule=true,
frametitlerulewidth=0.8pt,
frametitleaboveskip=0.3em,
frametitlebelowskip=0.5em}
\begin{mdframed}
\vspace{0.2em}

\strut \textbf{\small ChartQA Prompt}
\vspace{0.4em}
\hrule
\vspace{0.6em}

`Text': `Q1': [[q1]], `Q2': [[q2]]

\medskip

First solve Q1 and then solve Q2.

\medskip

Generate response in the format: The answer to Q1 is [answer]. The answer to Q2 is [answer].

\vspace{0.2em}
\end{mdframed}
\end{minipage}
\caption{MLLM generation prompt for ChartQA task.}
\label{fig:charta_generation_prompt}
\vspace{-1em}
\end{figure}
\begin{figure}[h!]
\centering
\noindent\begin{minipage}{\textwidth}
\mdfsetup{%
middlelinewidth=1pt,
backgroundcolor=yellow!7,
innerleftmargin=0.5cm,
innerrightmargin=0.5cm,
font=\small,
roundcorner=15pt,
frametitlebackgroundcolor=cyan!30,
frametitlefont=\bfseries\small,
frametitlerule=true,
frametitlerulewidth=0.8pt,
frametitleaboveskip=0.3em,
frametitlebelowskip=0.5em}
\begin{mdframed}
\vspace{0.2em}

\strut \textbf{\small VSR Prompt}
\vspace{0.4em}
\hrule
\vspace{0.6em}

`Text': [[caption]]

\medskip

There exists a discrepancy between the spatial location of key objects conveyed in the `Text' compared to the Image provided: [[instruction]].

\medskip

Identify this spatial discrepancy and answer in the format: the image has [object 1] [spatial\_relation] [object 2] while the text mentions [object 1] [spatial\_relation] [object 2].

\vspace{0.2em}
\end{mdframed}
\end{minipage}
\caption{MLLM generation prompt for VSR task.}
\label{fig:vsr_generation_prompt}
\vspace{-1em}
\end{figure}
\begin{figure}[h!]
\centering
\noindent\begin{minipage}{\textwidth}
\mdfsetup{%
middlelinewidth=1pt,
backgroundcolor=yellow!7,
innerleftmargin=0.5cm,
innerrightmargin=0.5cm,
font=\small,
roundcorner=15pt,
frametitlebackgroundcolor=cyan!30,
frametitlefont=\bfseries\small,
frametitlerule=true,
frametitlerulewidth=0.8pt,
frametitleaboveskip=0.3em,
frametitlebelowskip=0.5em}
\begin{mdframed}
\vspace{0.2em}
\strut \textbf{\small Math-Sport Prompt}
\vspace{0.4em}
\hrule
\vspace{0.6em}
`Math': [[sport\_text\_fill]]

\medskip

`Sport': [[math\_text\_fill]]

\medskip

First, solve the math problem in the 'Math' text and then identify the sport from the provided 'Sport' text.

\medskip

Generate response in the format: The answer to the math problem is [numeric\_answer]. The sport in the text is [sport].

\vspace{0.2em}
\end{mdframed}
\end{minipage}
\caption{MLLM generation prompt for Math-Sport task.}
\label{fig:generation_prompt_MaSp}
\vspace{-1em}
\end{figure}

We present the prompts (or instructions) given to the \lov{} and \qvl{} MLLMs across tasks.
The Fruit-Math prompt is shown in
\cref{fig:generation_prompt_FrMa},
ChartQA prompt in \cref{fig:charta_generation_prompt},
VSR prompt in \cref{fig:vsr_generation_prompt} and
the Math-Sport prompt fed to the base Qwen2-7B LLM is also included in \cref{fig:generation_prompt_MaSp}.

Note, the Fruit-Math prompt (text and instruction) is also illustrated separately in \cref{fig:teaser}A for one sample.

\subsection{Gemini POS Tagging Prompts}
\label{appendix:subsec:gemini_pos_prompts}

\begin{figure}[h!]
\centering
\noindent\begin{minipage}{\textwidth}
\mdfsetup{%
middlelinewidth=1pt,
backgroundcolor=yellow!7,
innerleftmargin=0.5cm,
innerrightmargin=0.5cm,
font=\small,
roundcorner=15pt}
\begin{mdframed}
\vspace{0.2em}
\textbf{Task:} You are a precise linguistic-semantic tagger.

Your job is to analyze a sequence of generated tokens and assign a semantic tag to each one.
\medskip

\textbf{Semantic Schema}\\
Here is the semantic schema you MUST use:
\begin{itemize}[leftmargin=*, nosep]
    \item[-] SOG: ``Start of Generation''. Use ONLY for the very first token.
    \item[-] FRUIT\_INTRO: ``Fruit Preamble''. Tokens leading up to the fruit name.
    \item[-] FRUIT\_CONCEPT: ``Fruit Entity''. The specific token(s) for the fruit name, including subwords.
    \item[-] HANDOFF: ``First Concept or Task Transition''. Tokens after the FRUIT\_CONCEPT. This token make up the shift from fruit to math solution, it neither belongs to fruit nor math solution.
    \item[-] MATH\_INTRO: ``Math Preamble''. Tokens leading up to the numeric math answer, including any symbols and special characters.
    \item[-] MATH\_ANSWER: ``Numerical Mathematical Answer''. The specific token(s) for the numerical answer to the puzzle, including decimals and delimiters.
    \item[-] POSTAMBLE: ``Conclusion''. Tokens after the final concept, including special characters, and any final punctuations.
\end{itemize}
\medskip

\textbf{Task}

Analyze the following list of tokens. Return a JSON list where each object in the list corresponds to a token and contains the ``word'' and its ``tag''.

The order of the tokens MUST be preserved in the list.

Respond ONLY with the JSON list.
\medskip

\textbf{Input Tokens:}
\begin{verbatim}
[[mllm_generated_response]]
\end{verbatim}
\vspace{0.2em}
\end{mdframed}
\end{minipage}
\caption{Gemini prompt for semantic POS token tagging for Fruit-Math task.}
\label{fig:semantic_tagging_prompt_FrMa}
\vspace{-1em}
\end{figure}
\begin{figure}[h!]
\centering
\noindent\begin{minipage}{\textwidth}
\mdfsetup{%
middlelinewidth=1pt,
backgroundcolor=yellow!7,
innerleftmargin=0.5cm,
innerrightmargin=0.5cm,
font=\small,
roundcorner=15pt}
\begin{mdframed}
\vspace{0.2em}

\textbf{Task:} You are a precise linguistic-semantic tagger.

Your job is to analyze a sequence of generated tokens and assign a semantic tag to each one.
\medskip

\textbf{Semantic Schema}\\
Here is the semantic schema you MUST use:

\begin{itemize}[leftmargin=*, nosep]

    \item[-] SOG: ``Start of Generation''. Use ONLY for the very first token.

    \item[-] IMG\_INTRO: Tokens leading up to the first object in the image description.

    \item[-] IMG\_OBJECT\_1: First object in the image.

    \item[-] IMG\_SPATIAL\_RELATION: Spatial relation in the image (e.g., ``left of'', ``above'').

    \item[-] IMG\_OBJECT\_2: Second object in the image.

    \item[-] HANDOFF: Tokens indicating contrast/transition between image and text (e.g., ``while'', ``but'', commas if structural).

    \item[-] TXT\_INTRO: Tokens leading up to the first object in the text description.

    \item[-] TXT\_OBJECT\_1: First object in the text.

    \item[-] TXT\_SPATIAL\_RELATION: Spatial relation in the text.

    \item[-] TXT\_OBJECT\_2: Second object in the text.

    \item[-] POSTAMBLE: Tokens after the final concept, including punctuation.

\end{itemize}
\medskip

\textbf{Task}

Analyze the following list of tokens. Return a JSON array where each object in the array corresponds to a token and contains the ``word'' and its ``tag''.

The order of the tokens MUST be preserved.

Respond ONLY with the JSON array.
\medskip

\textbf{Input Tokens:}
\begin{verbatim}
[[mllm_generated_response]]
\end{verbatim}

\vspace{0.2em}
\end{mdframed}
\end{minipage}

\caption{Gemini prompt for semantic POS token tagging for VSR task.}
\label{fig:semantic_tagging_prompt_vsr}

\vspace{-1em}
\end{figure}
\begin{figure}[h!]
\centering
\noindent\begin{minipage}{\textwidth}
\mdfsetup{%
middlelinewidth=1pt,
backgroundcolor=yellow!7,
innerleftmargin=0.5cm,
innerrightmargin=0.5cm,
font=\small,
roundcorner=15pt}
\begin{mdframed}
\vspace{0.2em}

\textbf{Task:} You are a precise linguistic-semantic tagger.

Your job is to analyze a sequence of generated tokens and assign a semantic tag to each one based on a visual Q\&A task.
\medskip

\textbf{Semantic Schema}\\
Here is the semantic schema you MUST use:
\begin{itemize}[leftmargin=*, nosep]
    \item[-] SOG: ``Start of Generation''. Use ONLY for the very first token (e.g., usually ``The'' or ``<pad>'').
    
    \item[-] Q1\_INTRO: ``Question 1 Preamble''. Tokens setting up the first answer (e.g., ``The'', ``answer'', ``to'', ``Q'', ``1'', ``is'').
    
    \item[-] Q1\_ANSWER: ``Question 1 Answer Entity''. The specific token(s) containing the extracted answer for the first question from the chart.
    
    \item[-] HANDOFF: ``Task Transition''. Tokens immediately following the Q1\_ANSWER (usually a punctuation mark like a comma).
    
    \item[-] Q2\_INTRO: ``Question 2 Preamble''. Tokens setting up the second answer (e.g., ``the'', ``answer'', ``to'', ``Q'', ``2'', ``is'').
    
    \item[-] Q2\_ANSWER: ``Question 2 Answer Entity''. The specific token(s) containing the extracted answer for the second question from the chart.
    
    \item[-] POSTAMBLE: ``Conclusion''. Tokens after the final answer concept, including final punctuations or End of Sequence tokens.
\end{itemize}
\medskip

\textbf{Task}

Analyze the following list of tokens. Return a JSON array where each object in the array corresponds to a token and contains the ``word'' and its ``tag''.

The order of the tokens MUST be preserved in the array.

Respond ONLY with the JSON array.
\medskip

\textbf{Input Tokens:}
\begin{verbatim}
[[mllm_generated_response]]
\end{verbatim}

\vspace{0.2em}
\end{mdframed}
\end{minipage}

\caption{Gemini prompt for semantic POS token tagging for the ChartQA task.}
\label{fig:semantic_tagging_prompt_chartqa}

\vspace{-1em}
\end{figure}

Next, we present the prompts used to convert variable-length outputs into a structured form through the `part-of-speech' like tags.
We present the prompts in
\cref{fig:semantic_tagging_prompt_FrMa} for Fruit-Math,
\cref{fig:semantic_tagging_prompt_vsr} for VSR and,
\cref{fig:semantic_tagging_prompt_chartqa} for ChartQA.

\begin{figure}[h!]
\centering
\noindent\begin{minipage}{\textwidth}
\mdfsetup{%
middlelinewidth=1pt,
backgroundcolor=yellow!7,
innerleftmargin=0.5cm,
innerrightmargin=0.5cm,
font=\small,
roundcorner=15pt}
\begin{mdframed}
\vspace{0.2em}
\textbf{Task:} You are a precise linguistic-semantic tagger.

Your job is to analyze a sequence of generated tokens and assign a semantic tag to each one.
\medskip

\textbf{Semantic Schema}\\
Here is the semantic schema you MUST use:
\begin{itemize}[leftmargin=*, nosep]
    \item[-] SOG: ``Start of Generation''. Use ONLY for the very first token.
    \item[-] MATH\_INTRO: ``Math Preamble''. Tokens leading up to the numeric math answer, including any symbols and special characters.
    \item[-] MATH\_ANSWER: ``Numerical Mathematical Answer''. The specific token(s) for the numerical answer to the puzzle, including decimals and delimiters.
    \item[-] FIRST\_HANDOFF: ``First Concept or Task Transition''. First token after the MATH\_ANSWER. This token make up the shift from math to sport solution, it neither belongs to math nor sport solution.
    \item[-] OTHER\_HANDOFF: ``Concept or Task Transition''. Token after the FIRST\_HANDOFF. These are tokens that make up the shift from math to sport solution, they neither belong to math nor sport solution.
    \item[-] SPORT\_INTRO: ``Sport Preamble''. Tokens leading up to the sport name.
    \item[-] SPORT\_CONCEPT: ``Sport Entity''. The specific token(s) for the sport name, including subwords.
    \item[-] POSTAMBLE: ``Conclusion''. Tokens after the final concept, including special characters, and any final punctuations.
\end{itemize}
\medskip

\textbf{Task}

Analyze the following list of tokens. Return a JSON list where each object in the list corresponds to a token and contains the ``word'' and its ``tag''.

The order of the tokens MUST be preserved in the list.

Respond ONLY with the JSON list.
\medskip

\textbf{Input Tokens:}
\begin{verbatim}
[[mllm_generated_response]]
\end{verbatim}
\vspace{0.2em}
\end{mdframed}
\end{minipage}
\caption{Gemini prompt for semantic POS token tagging for Math-Sport task.}
\label{fig:semantic_tagging_prompt_MaSp}
\vspace{-1em}
\end{figure}

The output of the LLM for the text-only task is also tagged using \cref{fig:semantic_tagging_prompt_MaSp}.

\subsection{Gemini Evaluation Prompts}
\label{appendix:subsec:gemini_eval_prompts}

\begin{figure}
\centering
\noindent\begin{minipage}{\textwidth}
\mdfsetup{%
middlelinewidth=1pt,
backgroundcolor=yellow!7,
innerleftmargin=0.5cm,
innerrightmargin=0.5cm,
font=\small,
roundcorner=15pt}
\begin{mdframed}
\vspace{0.2em}

\textbf{Task:} Evaluate the vision-language model response against the provided ground truth labels for fruit recognition and math reasoning.

\vspace{0.5em}

\textbf{Evaluation Objective}

You are an expert semantic evaluator. Grade the vision-language model response based on the ground truth.

\medskip

Be smart: give credit for synonyms (e.g., ``mandarin'' for ``orange'') and different formats (e.g., ``seven'' for ``7''). Focus on semantic intent.

\vspace{0.5em}

\textbf{Ground Truth}
\begin{itemize}[leftmargin=*, nosep]
    \item Expected Fruit: \texttt{[[gt\_fruit]]}
    \item Expected Math Answer: \texttt{[[gt\_math]]}
\end{itemize}

\vspace{0.5em}

\textbf{Model Response}
\begin{verbatim}
[[model_response]]
\end{verbatim}

\vspace{0.5em}

\textbf{Normalization Rules}
\begin{itemize}[leftmargin=*, nosep]
    \item[-] Fruit must be ONE canonical lowercase label (e.g., ``banana'').
    \item[-] If no fruit present $\rightarrow$ ``none''
    \item[-] If unclear $\rightarrow$ ``unknown''
    \item[-] Prefer the FINAL committed answer.
\end{itemize}

\vspace{0.5em}

\textbf{Evaluation Instructions}

\begin{enumerate}[leftmargin=*, nosep]
    \item Determine whether a fruit answer was provided.
    \item Determine whether the fruit answer is correct.
    \item Determine whether a math answer was provided.
    \item Determine whether the math answer is correct.
    \item Extract the final fruit prediction committed by the model.
    \item Extract the final numeric/math answer committed by the model.
    \item Copy the exact phrase corresponding to the final fruit answer and normalize it to a canonical label if possible.
    \item Copy the exact phrase corresponding to the final math answer and normalize it to a canonical form if possible.
    \item If the fruit prediction is incorrect, assign one error category:
    \begin{itemize}[leftmargin=*, nosep]
        \item[-] CROSS: Model is misdirected by terms/nouns from the math puzzle.
        \item[-] DENIAL: Model explicitly denies fruit presence or refuses to answer.
        \item[-] RESCUE: Model shows conflict or self-correction (e.g., ``I see no fruit... it is a cherry'' or ``Apple, no, Banana'').
        \item[-] PRIOR: Simple misidentification or guessing common fruits without evidence.
        \item[-] OTHER: None of the above. Briefly describe in \texttt{error\_explanation}.
        \item[-] NA: If fruit prediction is correct.
    \end{itemize}
    \item Provide a brief \texttt{error\_explanation}.
\end{enumerate}

\vspace{0.5em}

\textbf{Output Format (STRICT JSON)}

\begin{verbatim}
{
  "fruit_answered": "True/False",
  "fruit_correct": "True/False",
  "math_answered": "True/False",
  "math_correct": "True/False",
  "predicted_fruit": "string",
  "predicted_math": "string",
  "final_fruit_span": "string",
  "final_math_span": "string",
  "fruit_error_profiling": "CROSS/DENIAL/RESCUE/PRIOR/OTHER/NA",
  "error_explanation": "string"
}
\end{verbatim}

\vspace{0.2em}
\end{mdframed}
\end{minipage}
\caption{Accuracy, answer rate and error profiling prompt for the Fruit-Math task.}
\label{fig:accuracy_prompt_FrMa}
\vspace{-1em}
\end{figure}
\begin{figure}
\centering
\noindent\begin{minipage}{\textwidth}
\mdfsetup{%
middlelinewidth=1pt,
backgroundcolor=yellow!7,
innerleftmargin=0.5cm,
innerrightmargin=0.5cm,
font=\small,
roundcorner=15pt}
\begin{mdframed}
\vspace{0.2em}

\textbf{Task:} Evaluate whether the vision-language model correctly distinguishes mismatched spatial relations between the image and the accompanying text.

\vspace{0.5em}

\textbf{Evaluation Objective}

You are an expert evaluator grading a vision-language model's response.

\medskip

The model was presented with MISMATCHED visual and textual spatial relations.

\vspace{0.5em}

\textbf{Ground Truth}
\begin{itemize}[leftmargin=*, nosep]
    \item Expected Image Description (Image): \texttt{\{expected\_image\_relation\}}
    \item Expected Text Description (Text): \texttt{\{expected\_text\_relation\}}
\end{itemize}

\vspace{0.5em}

\textbf{Model Response}
\begin{verbatim}
{model_response}
\end{verbatim}

\vspace{0.5em}

\textbf{Evaluation Instructions}

\begin{enumerate}[leftmargin=*, nosep]
    \item Visual Check: Determine whether the model correctly identifies \texttt{\{expected\_image\_relation\}} as the spatial relation present in the image. Output ``True'' or ``False''.
    
    \item Text Check: Determine whether the model correctly identifies \texttt{\{expected\_text\_relation\}} as the spatial relation described in the text. Output ``True'' or ``False''.
    
    \item Alignment Classification:
    \begin{itemize}[leftmargin=*, nosep]
        \item[-] \texttt{correct}: Both image and text relations are correctly identified.
        
        \item[-] \texttt{incorrect - both image}: Model claims both modalities show or indicate the image relation.
        
        \item[-] \texttt{incorrect - both text}: Model claims both modalities show or indicate the text relation.
        
        \item[-] \texttt{incorrect - flip}: Model swaps them, using the text relation for the image part and the image relation for the text part.
        
        \item[-] \texttt{incorrect - other}: Hallucination, refusal, incomplete response, or nonsensical output.
    \end{itemize}
\end{enumerate}

\vspace{0.5em}

\textbf{Output Format (STRICT JSON)}

\begin{verbatim}
{
  "visual_check": "True/False",
  "text_check": "True/False",
  "alignment_classification":
    correct
    incorrect - both image
    incorrect - both text
    incorrect - flip
    incorrect - other"
}
\end{verbatim}

\vspace{0.2em}
\end{mdframed}
\end{minipage}
\caption{Evaluation prompt for the VSR task.}
\label{fig:acc_vsr_prompt}
\vspace{-1em}
\end{figure}

Finally, the accuracy evaluation of generated outputs (fruit and math accuracy) is also achieved through Gemini.
\cref{fig:accuracy_prompt_FrMa} presents the Fruit-Math prompt and \cref{fig:acc_vsr_prompt} shows the VSR evaluation prompt.
We manually verify the outputs to ensure robustness of the LLM-based evaluation.

\section{Impact Statement}
\label{appendix:subsec:impact_statement}
This paper presents work that aims to advance the interpretability of Multimodal LLMs.
This is an active area of research, and LLMs and MLLMs are increasingly being adopted at scale.
While there are potential societal consequences of LLM and MLLM usage in general, we feel that none need to be specifically highlighted here with respect to this particular work on analyzing how attention patterns evolve.

\section{Compute Resources}
\label{appendix:subsec:compute}
Our work is a training-free analysis and intervention effort, and does not require a large-scale compute setup.
All our experiments are conducted on the NVIDIA A6000 GPUs with 48GB VRAM.


\end{document}